\documentclass{article} 
\usepackage{iclr2027_conference,times}

\usepackage{amsmath,amsfonts,bm}

\def\eqref#1{equation~\ref{#1}}

\def\1{\bm{1}}

\DeclareMathAlphabet{\mathsfit}{\encodingdefault}{\sfdefault}{m}{sl}
\SetMathAlphabet{\mathsfit}{bold}{\encodingdefault}{\sfdefault}{bx}{n}

\usepackage{hyperref}
\usepackage{url}
\usepackage{booktabs}       
\usepackage{amsfonts}       
\usepackage{nicefrac}       
\usepackage{xcolor}         
\usepackage{multirow}
\usepackage{tabularx}
\usepackage{colortbl}
\usepackage{arydshln}
\usepackage{wrapfig}
\usepackage{amsmath}
\usepackage{stfloats}
\usepackage{graphicx}

\title{Boosting Large Language Models with Mask Fine-Tuning}

\author{
\begin{tabular}{@{}l@{}}
\textbf{Mingyuan Zhang$^{1}$, Yue Bai$^{1}$, Huan Wang$^{1}$, Yizhou Wang$^{1}$, Qihua Dong$^{1}$,}\\
\textbf{Yitian Zhang$^{1}$, Yun Fu$^{1,2}$}
\end{tabular}\\[4pt]
$^{1}$Department of Electrical and Computer Engineering, Northeastern University\\
$^{2}$Khoury College of Computer Science, Northeastern University
}

\iclrfinalcopy 
\begin{document}

\maketitle
\lhead{}
\renewcommand{\headrulewidth}{0pt}

\begin{abstract}
The large language model (LLM) is typically integrated into the mainstream optimization protocol.
However, it remains underexplored whether maintaining the model integrity is \textit{indispensable} for promising performance.
In this work, we introduce Mask Fine-Tuning (MFT), a novel LLM fine-tuning paradigm demonstrating that carefully breaking the model's structural integrity can surprisingly improve performance without updating model weights.
MFT learns and applies binary masks to well-optimized models, using the standard LLM fine-tuning objective as supervision.
Based on fully fine-tuned models, MFT uses the same fine-tuning datasets to achieve consistent performance gains across domains and backbones (e.g., an average gain of \textbf{2.70/4.15} on IFEval with LLaMA2-7B/3.1-8B).
Detailed ablation studies and analyses examine the proposed MFT from different perspectives, including the sparse ratio and the loss surface.
Additionally, when deployed on well-trained models, MFT is compatible with other LLM optimization procedures to improve overall model performance.
Furthermore, this study extends the masking operation beyond its conventional use in network pruning for model compression to encompass a broader range of model capabilities. Codes: \emph{\href{https://github.com/Ming-K9/MFT}
{\textcolor{magenta}{https://github.com/Ming-K9/MFT}}}.
\end{abstract}

\section{Introduction}\label{sec:intro}
Pre-training large language models (LLMs) typically requires massive corpora to enable models with foundational knowledge and linguistic competencies for effective language generation~\citep{radford2019language, brown2020language, touvron2023llama, dubey2024llama}.
Afterward, researchers fine-tune pre-trained LLMs for downstream tasks using high-quality domain data, including specialized knowledge such as math and coding~\citep{cobbe2021training, yu2023metamath, chen2021evaluating, austin2021program} and specific patterns such as human instructions~\citep {zhou2023instruction, dubois2024length}.
For such adaptations, fine-tuning all model parameters is the most straightforward and common approach, often called full fine-tuning (FFT), and generally yields the most competitive performance.
Furthermore, adapter-based alternatives offer parameter-efficient fine-tuning (PEFT) strategies such as vanilla bottleneck adapters~\citep{houlsby2019parameter} and LoRA~\citep{hu2021lora}.
These methods freeze the pre-trained backbone and add a limited number of learnable parameters.
Although these methods may sacrifice fine-tuning performance relative to FFT, they suit scenarios with limited computational budgets, especially when fine-tuning data are insufficient.
Overall, the pipeline of language model pre-training followed by fine-tuning has demonstrated promising results across a range of language tasks, including domains requiring extensive specialized knowledge (e.g., medicine~\citep{wang2023clinicalgpt} and law~\citep{nguyen2023brief}) and complex reasoning (e.g., math~\citep{frieder2024mathematical} and coding~\citep{chen2021evaluating}), exhibiting remarkably intelligent capabilities.

\begin{figure*}[t]
    \centering
    \includegraphics[width=1\linewidth]{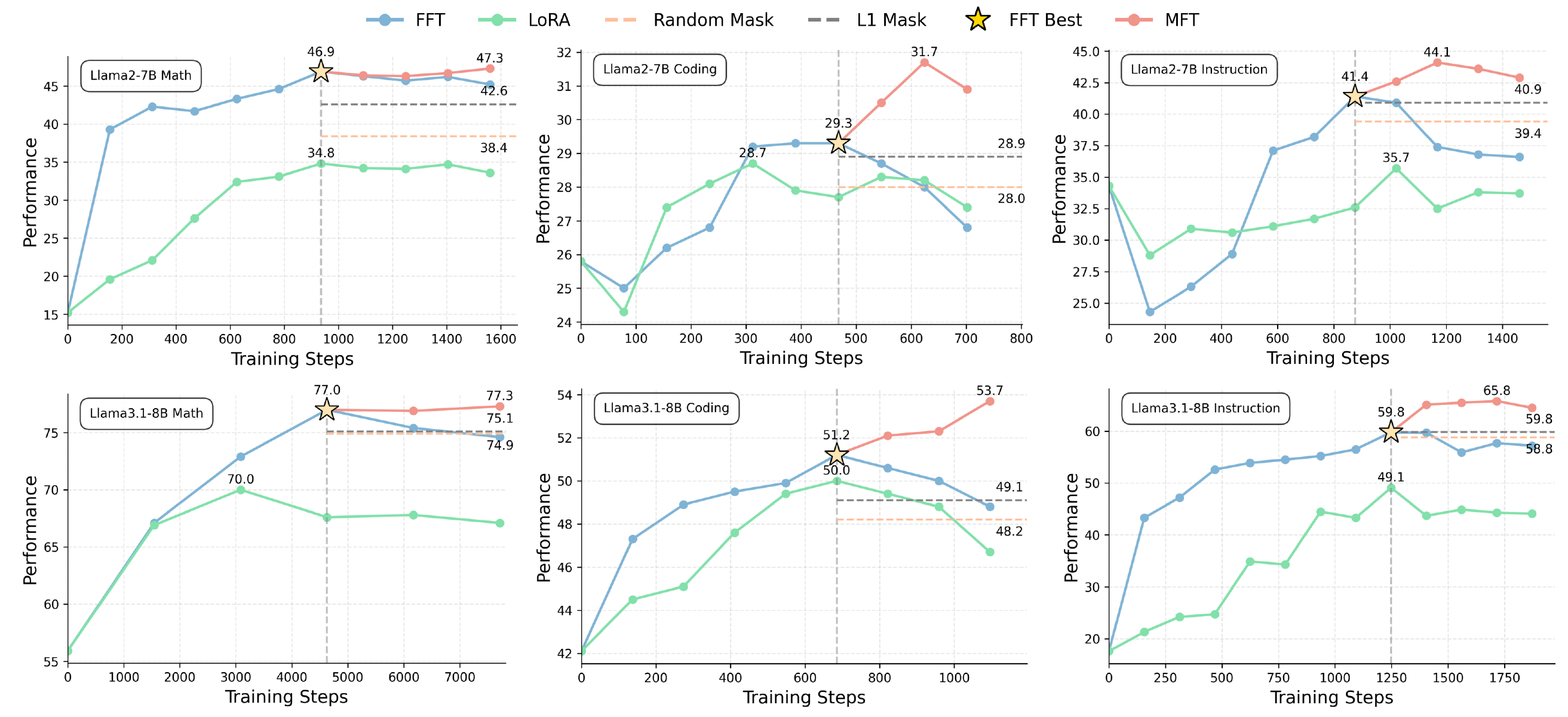}
    \caption{
    The visualization of the performance trend across different fine-tuning strategies, including FFT (blue line), LoRA (green line)~\citep{hu2021lora}, and our MFT (red line).
    We also add random (orange dashed) and L1 (gray dashed) masks for comparison.
    We use three settings across LLaMA and Qwen backbones on GSM8K, HumanEval, and IF-Eval for math, coding, and instruction domains, respectively.
    The x-axis is training steps starting from the pre-trained backbone.
    The y-axis is evaluation performance.
    MFT (red line) starts from the best FFT model (yellow star) and improves upon the upper bound, whereas continued FFT leads to overfitting.
    It also outperforms LoRA fine-tuning and two vanilla mask baselines.
    }
    \label{fig:intro_teaser}
\end{figure*}

However, the mainstream LLM optimization process (pre-training followed by fine-tuning), which is promisingly used to benchmark modern LLM performance, consistently treats the language model as a whole.
For pretraining, the model structure remains mostly dense.
For fine-tuning, FFT optimizes all model parameters simultaneously, while PEFT freezes them to tune additional adapter parameters, with both approaches accepting the necessity of LLM structural integrity by default.
This practice naturally leads us to ask: \textit{Is such structural integrity indispensable for good model performance? Or, more broadly, can we improve the model by removing certain components that break structural integrity?}

We propose \textit{mask fine-tuning (MFT)} to answer this question.
MFT freezes a given LLM model and learns a binary mask over it to assess whether breaking structural integrity works.
To substantiate its effectiveness, we use a well-trained LLM (the model after sufficient FFT in this study) as a strong baseline to assess whether MFT improves further.
MFT uses the same standard LLM fine-tuning objective and datasets as FFT to learn the mask.
The only difference is that the learnable part shifts from model weights to a binary weight mask, while the weights themselves remain fixed.
The learned mask indicates which parameter positions to remove. 
Surprisingly, we find that masking these weights yields further performance gains compared with the given well-trained model.
As shown in Fig.~\ref{fig:intro_teaser}, the regular FFT (blue line) improves the pre-trained model across settings, but once it reaches peak performance (yellow star), excessive FFT leads to overfitting.
However, MFT (red line) starts from the best FFT model and improves it further.
These teaser cases show a consistent observation across different backbones (LLaMA2-7B~\citep{touvron2023llama} and LLaMA3.1-8B~\citep{dubey2024llama}) and domains (GSM8K~\citep{cobbe2021training}, HumanEval~\citep{chen2021evaluating}, and IF-Eval~\citep{zhou2023instruction} for math, coding, and instruction, respectively).
We elaborate comparisons of more configurations in Sec.~\ref{sec:exps}.
This phenomenon substantially answers the question we are interested in: \textit{LLM structural integrity is not indispensable for good model performance, and breaking it can lead to further improvements}.

Here, we primarily explore MFT as a post-fine-tuning strategy for further improvement, starting from an FFT checkpoint that yet outperforms.
It naturally upgrades the current fine-tuning routine to a new protocol, as shown in Fig.~\ref{fig:intro_pipeline}, offering a new perspective on investigating LLM fine-tuning.
Furthermore, MFT is compatible with existing fine-tuning methods and can be integrated flexibly into any of them, as it shares the same optimization objective and requires no additional data annotation.

In addition, we highlight the difference between typical pruning methods and our work.
The former compresses the model while preserving its trained capabilities, whereas the latter seeks further improvement beyond a well-trained model for a specific purpose.
Even if they share the same concept of \textit{mask} (or model sparsity), their fundamental goals are different (see more discussions in Sec.~\ref {sec:literature}).
Along these lines, our study extends the masking operation beyond typical network compression to a more general model-capability setting.
In other words, analogically speaking, \textit{typical masking employs subtraction to reduce (pruning to compress), whereas our approach leverages subtraction to achieve augmentation (removing weights to improve).}
Our contributions are as follows:
\begin{itemize}
    \item We validate that a well-trained LLM can be further improved by carefully removing certain weights using the same objective and dataset as regular fine-tuning, with limited computational overhead.
    This strategy is compatible with other training pipelines and improves current LLM optimization.
    
    \item We propose \textit{Mask Fine-Tuning (MFT)} to complete our exploration in a post fine-tuning scenario.
    MFT starts from a competitive, fully fine-tuned model with fixed weights and learns a binary mask that is applied to the model to improve it.
    Following this, MFT treats model sparsity in a new light, not only for efficiency, but also for performance improvement.
    
    \item Extensive experiments across different backbones (LLaMA and Qwen), domains (math, coding, and instruction), and FFT settings (domain-specific and mixed-up) show MFT's effectiveness, with consistent performance gains.
    Detailed ablations and analyses provide better intuition and inspire future work.
\end{itemize}

\section{Mask Fine-Tuning}\label{sec:method}
\begin{wrapfigure}{r}{0.52\columnwidth}
    \centering
    \vspace{-16pt}
    \includegraphics[width=\linewidth]{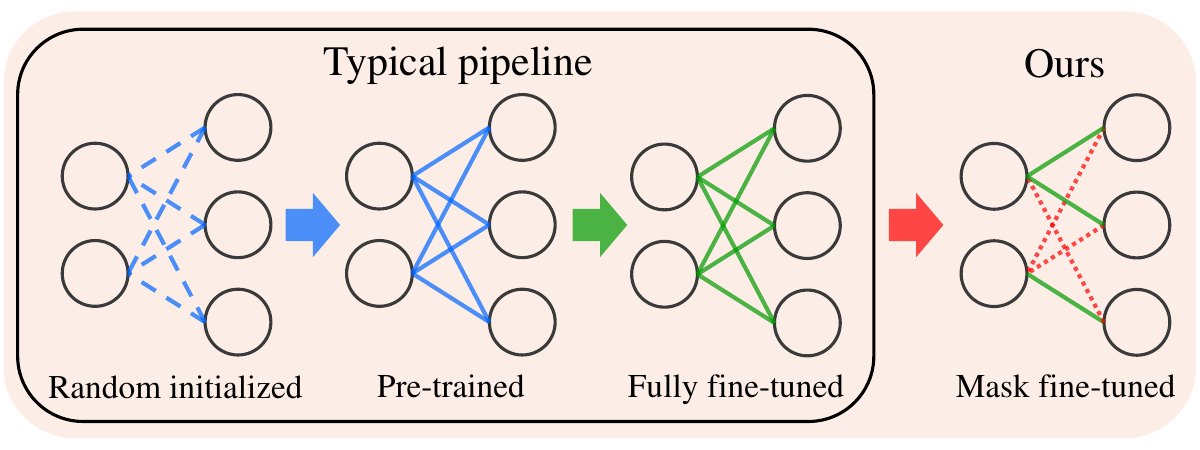}
    \vspace{-8mm}
    \caption{
    Typical LLM training comprises pre-training and fine-tuning to build capacity and domain knowledge, with the structure always being the entire model.
    We ask whether this integrity is necessary and propose that MFT generally outperforms models with sufficient FFT, naturally upgrading the classic pipeline by following the typical protocol to further refine well-optimized LLMs.
    }
    \label{fig:intro_pipeline}
    \vspace{-16pt}
\end{wrapfigure}
We focus on GPT-like language models~\citep{brown2020language,radford2019language} in an autoregressive manner and begin with a fully fine-tuned model built on a pre-trained language backbone for masked fine-tuning (MFT).
We briefly introduce the language model notation for full fine-tuning (FFT), then present our MFT.

\noindent \textbf{Full Fine-Tuning.}
We refer a pre-trained auto-regressive language backbone as $\mathcal{N}_p$ with parameters $\Theta_p$ and optimize the following objective to represent a FFT:
\begin{equation}\label{eq:fft}
L(U_f) = \sum_i log P(u_f^i | u_f^{i-k},...,u_f^{i-1};\Theta_p),
\end{equation}
where $L$ is the autoregressive loss used to supervise next-token prediction.
$\Theta_p$ is fully optimized, and $U_f = \{u_f^1, \ dots,u_f^n\}$ is a token sequence from a language corpus, serving as an FFT dataset.
We denote the model after sufficient FFT as $\mathcal{N}_f$, and its optimized parameters as $\Theta_f$.

\noindent \textbf{Mask Fine-Tuning.}
We deploy our MFT on models after the FFT.
Formally, we revise Eq.~\eqref{eq:fft} by adding a binary mask onto model parameters and obtain the rewritten loss for MFT as below:
\begin{equation}\label{eq:mft}
L(U_m) = \sum_i log P(u_m^i | u_m^{i-k},...,u_m^{i-1}; \Theta_f \odot M),
\end{equation}
where it shares the same objective function $L$ as in FFT and uses the dataset $U_m$ for supervision.
In our experiments, the dataset $U_m$ for MFT is the same as $U_f$ for FFT.
The main difference is that we add a binary mask $M$ to the model parameters $\Theta_f$. $M$ and $\Theta_f$ have the same size, and we mask out the corresponding weights in $\Theta_f$ via element-wise multiplication $\odot$.
During MFT, $\Theta_f$ is fixed, and $M$ is updated.
This way, MFT learns to locate specific weights while keeping the fully fine-tuned parameters $\Theta_f$ unchanged.
We can remove these weights from the well-trained model by applying the learned binary mask.
We optimize $M$ using a learnable process based on the straight-through gradient estimator~\citep{bengio2013estimating}, which we call the learnable mask.

\noindent \textbf{Learnable Mask.}
Given a model parameter $\Theta_f$ after FFT, we represent it layer-wise as $\Theta = \{\theta_l\}, l \in \{1,2,...,L\}$, omitting the subscript $f$ for notational simplicity.
$L$ denotes the number of model layers.
Then, we can formalize each layer as
\begin{equation}\label{eq:ste_1}
I_{l+1} = \sigma (F [I_l;\theta_l]),
\end{equation}
where $I_l$ is the layer $l$ input and $I_{l+1}$ is its output with activation $\sigma$.
$F$ generally serves as a layer operation (e.g., a convolutional or linear layer) with a parameter
$\theta_l = \{\theta_l^d\}, d \in \{1,2,...,D_l\}$.
$D_l$ represents the layer $l$ parameter dimension.
In this work, $F$ denotes a linear layer, as we primarily consider Transformer-based language models and apply the learnable mask to the model's linear mappings.
To learn the mask, we fix all parameters $\Theta$, denoted $\overline{\Theta}$.
Then, we assign each weight $\overline {\theta_l^d}$ a learnable score $c_l^d$, representing the importance of the corresponding weight.
Based on this, we rewrite Eq.~\eqref{eq:ste_1} as
\begin{equation}\label{eq:ste_2}
I_{l+1} = \sigma (F [I_l;\overline {\theta_l} \odot v (c_l)]),
\end{equation}
where $c_l = \{c_l^1, c_l^2, ..., c_l^{D_l}\}$ is the score of parameter $\overline {\theta_l}$.
$v$ is an indicator function to select a mask based on scores.
In this work, we use a ratio-based version in which $v$ outputs 1 if $c_l^d$ is among the top $K\%$ highest-scoring values, and 0 otherwise, where $K$ is a pre-defined mask ratio.
By updating $c$ with fixed $\overline\Theta$, we keep some parameters in $\overline\Theta$ while masking out others.
As the indicator function $v$ is non-differentiable, we use the straight-through gradient estimator~\citep{bengio2013estimating} to estimate the gradient of the loss with respect to $c_l^d$.
Concretely, we treat $v$ as an identity function during the backward pass. Then, the approximation can be described as
\begin{equation}\label{eq:ste_3}
\tilde{g}(c_l^d) = \frac{\partial L}{\partial \tilde I_{l+1}} \frac {\partial \tilde I_{l+1}} {\partial c_l^d} \approx \frac{\partial L}{\partial I_{l+1}} \frac {\partial I_{l+1}} {\partial c_l^d},
\end{equation}
where $\tilde I_{l+1} = \sigma (F [I_l; \overline \theta_l \odot c_l])$ after applying estimation and $\tilde g(c_l^d)$ is the estimated gradient with respect to score $c_l^d$.
In this way, MFT learns to obtain a mask for the model $\mathcal{N}_f$, and we denote the resulting model as $\mathcal{N}_m$ with parameters $\Theta_f$ and mask $M$.

MFT learns to identify specific weights using the regular training loss of language models, then removes them by applying the learned mask.
It answers our question (Sec.~\ref{sec:intro}) and introduces a new LLM fine-tuning protocol.
We illustrate the MFT procedure in Fig.~\ref{fig:intro_pipeline}.
Please note that the FFT method used in this study is supervised fine-tuning (SFT) and does not consider other policy-based tuning methods like DPO (Direct Preference Optimization) ~\citep{rafailov2024direct} or PPO (Proximal Policy Optimization)~\citep{schulman2017proximal}.
Our MFT generalizes easily to them. We leave such explorations to our future work.

\section{Experiments}\label{sec:exps}
We evaluate the proposed mask fine-tuning (MFT) on large language models (LLMs) across several backbones, domains, and full fine-tuning (FFT) settings.
In addition, we provide ablation studies and visualizations to provide better intuition.

\subsection{Experimental Setups}\label{sec:exps_pre}
\noindent \textbf{Pretrained Backbones.}
We use Transformer~\citep{vaswani2017attention} based pretrained LLMs as backbone models, including LLaMA2-7B~\citep{touvron2023llama}, LLaMA3.1-8B~\citep{dubey2024llama}, Qwen3-1.7B~\citep{yang2025qwen3technicalreport}, and Qwen3-14B~\citep{yang2025qwen3technicalreport}.

\noindent \textbf{Domains \& Datasets.}
We involve three domains, each with its representative tasks.
Specifically, we include GSM8K~\citep{cobbe2021training} and MetaMath~\citep{yu2023metamath} for math, HumanEval and HumanEval+~\citep{chen2021evaluating} for coding, and IF-Eval~\citep{zhou2023instruction} and Alpaca-Eval~\citep{dubois2024length} for instruction following domains, respectively.
We use all three domains for typical FFT, our MFT, and evaluation.

\begin{figure*}[t]
    \centering
    \includegraphics[width=1\linewidth]{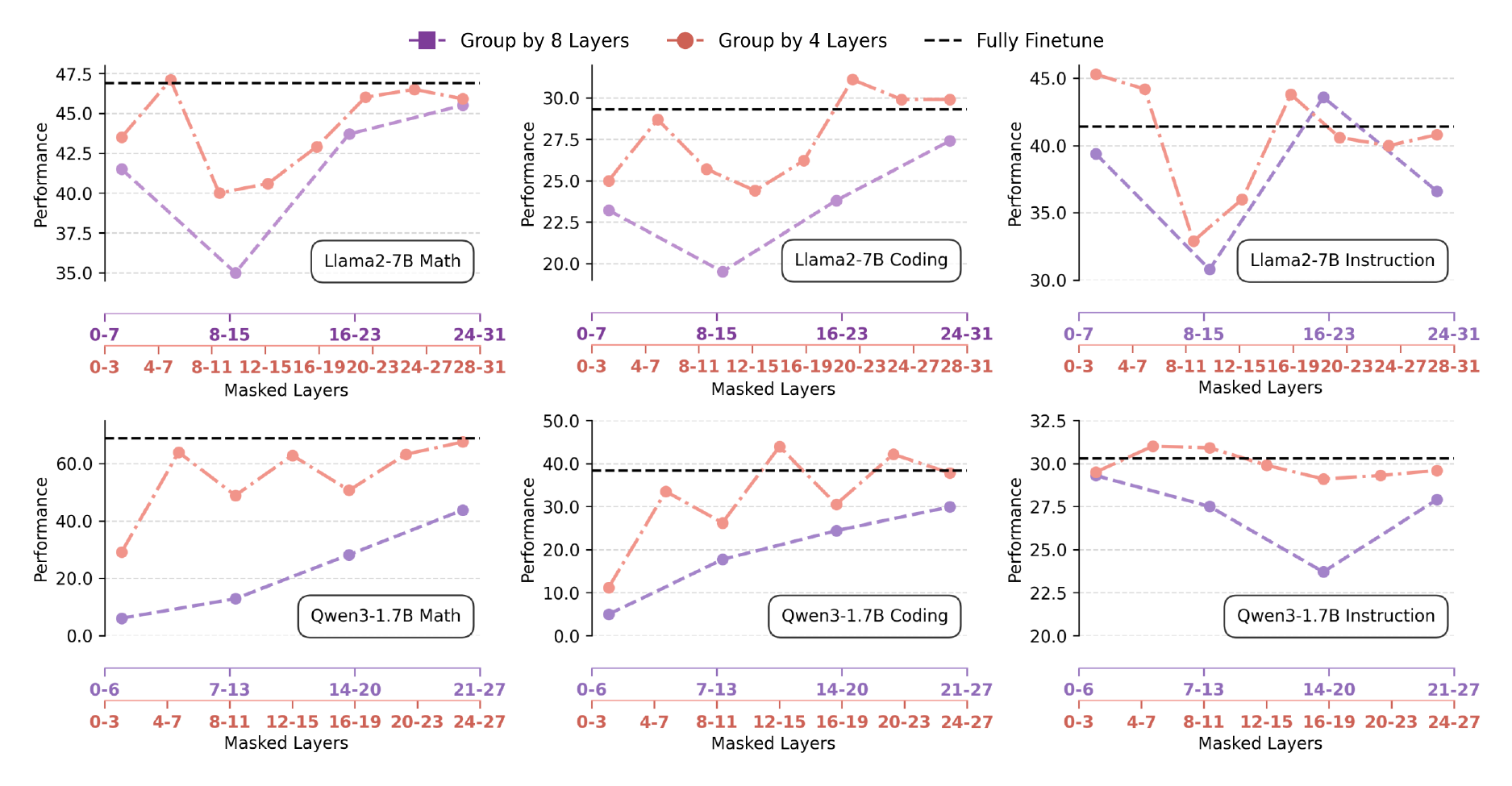}
    \caption{Visualization of the ablation study of the local MFT strategy.
    It uses LLaMA2-7B and Qwen3-1.7B backbones, covering math, coding, and instruction domains.
    In each figure, we conduct MFT ablations with a 10\% masking ratio on a domain-specific FFT model (black dashed line).
    We swap the ablation between two local granularities, 8/7-layer (purple) and 4-layer (orange), moving from shallow to deep layers using a 10\% fine-tuning set.
    This ablation is intended for quick intuition, with limited performance gains from MFT on a 10\% subset.
    Based on this trend, we deploy MFT with complete fine-tuning sets, achieving more improvements.
    }
    \label{fig:exps_local_ablation}
\end{figure*}

\noindent \textbf{Configurations.}
Our experiments use pretrained backbones, with FFT, MFT, and downstream evaluation in that order.
For FFT, we follow two strategies: 1) domain-specific, 2) mix-up, where the former conducts FFT using datasets from individual domains, such as math, coding, or instruction following, and the latter uses a mix-up dataset from all these domains.
For MFT, we follow a domain-specific approach and use datasets from a single domain.
Further details of the dataset are provided in the Appendix.
We mainly investigate local masking in this paper, as elaborated in a series of proof-of-concept studies in Sec.~\ref{sec:exps_main_results} and Fig.~\ref{fig:exps_local_ablation}.
We also provide an initial exploration of global masking strategy in Sec.~\ref{sec:exps_ablation} and Tab.~\ref{tab:exps_global_mask}.
For evaluation, we comprehensively assess performance changes (see Tab.~\ref{tab:exps_main}).

\noindent \textbf{Baselines.}
We use several baselines for comparison, including (1) a very basic pre-trained backbone;
(2) FFT as a strong baseline, which is mostly seen as the most competitive approach for downstream fine-tuning.
We perform a sufficient number of FFTs and report the best performance;
(3) LoRA~\citep{hu2021lora} fine-tuning as a strong alternative of FFT;
(4) continued FFT, which always causes a performance drop due to overfitting;
(5) continued LoRA plays a similar role for the LoRA baseline.
We train all baselines for 4 epochs to ensure sufficient training and report the final performance for Continued FFT and Continued LoRA.
(6) random mask and (7) L1 mask as two vanilla masking strategies.
Among them, (2) FFT and (4) continued FFT are critical to our experiments, since our goal is to use MFT to further improve the best FFT, and continued FFT serves as a sanity check.

\begin{table*}[t]
    \caption{
    Performance comparison of domain-specific settings.
    Each part has three blocks containing 1) common LoRA and its continued variant (green), 2) FFT serving as an upper bound and its continued variant (blue), and 3) our MFT (red) with the other two vanilla masking baselines.
    The pre-trained performance serves as a lower bound.
    }
    \centering
    \renewcommand{\arraystretch}{1.25}
    \setlength{\tabcolsep}{12pt}
    \resizebox{\textwidth}{!}{
    \begin{tabular}{l|lllllll}
        \hline
        \multicolumn{2}{c}{\multirow{2}{*}{\textbf{Method}}} 
        & \multicolumn{2}{c}{\textbf{Math}} 
        & \multicolumn{2}{c}{\textbf{Coding}} 
        & \multicolumn{2}{c}{\textbf{Instruction Following}} \\
        \cline{3-8}
        \multicolumn{2}{l}{} 
        & \multicolumn{1}{c}{GSM8K} 
        & \multicolumn{1}{c}{Math} 
        & \multicolumn{1}{c}{HumanEval} 
        & \multicolumn{1}{c}{HumanEval+} 
        & \multicolumn{1}{c}{IF-Eval} 
        & \multicolumn{1}{c}{Alpaca-Eval} \\
        \hline
        \multirow{8}{*}{\rotatebox[origin=c]{90}{\textbf{LLaMA2-7B}}}
        & \textbf{Pre-Trained Model}
        & 15.2 
        & 2.5 
        & 25.8 
        & 22.4 
        & 34.3 
        & 0.5 \\
        & Best LoRA \cellcolor[HTML]{EAFBF0} 
        & $34.8_{\pm0.34}$ \cellcolor[HTML]{EAFBF0} 
        & $4.7_{\pm0.26}$ \cellcolor[HTML]{EAFBF0} 
        & $28.7_{\pm0.56}$ \cellcolor[HTML]{EAFBF0} 
        & $23.8_{\pm0.48}$ \cellcolor[HTML]{EAFBF0} 
        & $35.7_{\pm0.68}$ \cellcolor[HTML]{EAFBF0} 
        & $1.2_{\pm0.16}$ \cellcolor[HTML]{EAFBF0} \\
        & Continued LoRA \cellcolor[HTML]{EAFBF0} 
        & $33.6_{\pm0.42}$ \cellcolor[HTML]{EAFBF0} 
        & $4.5_{\pm0.23}$ \cellcolor[HTML]{EAFBF0} 
        & $28.2_{\pm0.49}$ \cellcolor[HTML]{EAFBF0} 
        & $23.2_{\pm0.43}$ \cellcolor[HTML]{EAFBF0} 
        & $32.5_{\pm0.84}$ \cellcolor[HTML]{EAFBF0} 
        & $1.1_{\pm0.12}$ \cellcolor[HTML]{EAFBF0} \\
        & \textbf{Best FFT} \cellcolor[HTML]{E6F2FB} 
        & $46.8_{\pm0.22}$ \cellcolor[HTML]{E6F2FB} 
        & $6.6_{\pm0.21}$ \cellcolor[HTML]{E6F2FB} 
        & $29.4_{\pm0.36}$ \cellcolor[HTML]{E6F2FB} 
        & $25.1_{\pm0.34}$ \cellcolor[HTML]{E6F2FB} 
        & $41.2_{\pm0.62}$ \cellcolor[HTML]{E6F2FB} 
        & $1.9_{\pm0.18}$ \cellcolor[HTML]{E6F2FB} \\
        & {Continued FFT} \cellcolor[HTML]{E6F2FB} 
        & $45.0_{\pm0.28}$ \textbf{\textcolor[rgb]{0.75,0.0,0.0}{↓}} \scriptsize{1.8} \cellcolor[HTML]{E6F2FB}
        & $5.5_{\pm0.23}$ \textbf{\textcolor[rgb]{0.75,0.0,0.0}{↓}} \scriptsize{1.1} \cellcolor[HTML]{E6F2FB}
        & $27.9_{\pm0.42}$ \textbf{\textcolor[rgb]{0.75,0.0,0.0}{↓}} \scriptsize{1.5} \cellcolor[HTML]{E6F2FB}
        & $23.6_{\pm0.38}$ \textbf{\textcolor[rgb]{0.75,0.0,0.0}{↓}} \scriptsize{1.5} \cellcolor[HTML]{E6F2FB}
        & $37.8_{\pm0.68}$ \textbf{\textcolor[rgb]{0.75,0.0,0.0}{↓}} \scriptsize{3.4} \cellcolor[HTML]{E6F2FB}
        & $2.0_{\pm0.21}$ \textbf{\textcolor[rgb]{0.0, 0.75, 0.0}{↑}} \scriptsize{0.1} \cellcolor[HTML]{E6F2FB}\\
        & {Random Mask \scriptsize{w/ Best FFT}} 
        & $38.4_{\pm0.38}$ 
        & $4.9_{\pm0.25}$ 
        & $28.0_{\pm0.51}$ 
        & $24.0_{\pm0.47}$ 
        & $39.4_{\pm0.99}$ 
        & $1.4_{\pm0.21}$ \\
        & {L1 Mask \scriptsize{w/ Best FFT}} 
        & 42.6 
        & 5.7 
        & 28.9 
        & 24.5 
        & 40.9 
        & 1.5 \\
        & {MFT \scriptsize{w/ Best FFT}} \textbf{(Ours)} \cellcolor[HTML]{FCE7E4}
        & $\textbf{47.3}_{\pm0.19}$ \textbf{\textcolor[rgb]{0.0, 0.75, 0.0}{↑}} \scriptsize{0.5}  \cellcolor[HTML]{FCE7E4}
        & $\textbf{7.4}_{\pm0.24}$ \textbf{\textcolor[rgb]{0.0, 0.75, 0.0}{↑}} \scriptsize{0.8}  \cellcolor[HTML]{FCE7E4}
        & $\textbf{31.8}_{\pm0.31}$ \textbf{\textcolor[rgb]{0.0, 0.75, 0.0}{↑}} \scriptsize{2.4}  \cellcolor[HTML]{FCE7E4}
        & $\textbf{27.9}_{\pm0.25}$ \textbf{\textcolor[rgb]{0.0, 0.75, 0.0}{↑}} \scriptsize{2.8}  \cellcolor[HTML]{FCE7E4}
        & $\textbf{44.1}_{\pm0.72}$ \textbf{\textcolor[rgb]{0.0, 0.75, 0.0}{↑}} \scriptsize{2.9}  \cellcolor[HTML]{FCE7E4}
        & $\textbf{3.0}_{\pm0.22}$ \textbf{\textcolor[rgb]{0.0, 0.75, 0.0}{↑}} \scriptsize{1.1}  \cellcolor[HTML]{FCE7E4} \\
        \hline
        \multirow{8}{*}{\rotatebox[origin=c]{90}{\textbf{LLaMA3.1-8B}}} 
        & \textbf{Pre-Trained Model}
        & 55.9 
        & 14.6 
        & 42.1 
        & 37.8 
        & 17.6 
        & 9.0 \\
        & Best LoRA \cellcolor[HTML]{EAFBF0} 
        & $70.0_{\pm0.28}$  \cellcolor[HTML]{EAFBF0} 
        & $22.7_{\pm0.22}$  \cellcolor[HTML]{EAFBF0} 
        & $50.0_{\pm0.32}$  \cellcolor[HTML]{EAFBF0} 
        & $45.1_{\pm0.43}$  \cellcolor[HTML]{EAFBF0} 
        & $49.1_{\pm0.92}$  \cellcolor[HTML]{EAFBF0} 
        & $10.1_{\pm0.51}$  \cellcolor[HTML]{EAFBF0} \\
        & Continued LoRA \cellcolor[HTML]{EAFBF0} 
        & $67.1_{\pm0.35}$  \cellcolor[HTML]{EAFBF0} 
        & $19.7_{\pm0.19}$  \cellcolor[HTML]{EAFBF0} 
        & $46.7_{\pm0.46}$  \cellcolor[HTML]{EAFBF0} 
        & $44.5_{\pm0.38}$  \cellcolor[HTML]{EAFBF0} 
        & $44.3_{\pm0.88}$  \cellcolor[HTML]{EAFBF0} 
        & $8.9_{\pm0.29}$  \cellcolor[HTML]{EAFBF0} \\
        & {\textbf{Best FFT}} \cellcolor[HTML]{E6F2FB} 
        & $76.8_{\pm0.18}$  \cellcolor[HTML]{E6F2FB} 
        & $24.2_{\pm0.18}$  \cellcolor[HTML]{E6F2FB} 
        & $51.0_{\pm0.39}$  \cellcolor[HTML]{E6F2FB} 
        & $44.9_{\pm0.31}$  \cellcolor[HTML]{E6F2FB} 
        & $59.6_{\pm0.69}$  \cellcolor[HTML]{E6F2FB} 
        & $11.9_{\pm0.36}$  \cellcolor[HTML]{E6F2FB} \\
        & {Continued FFT} \cellcolor[HTML]{E6F2FB} 
        & $74.4_{\pm0.17}$ \textbf{\textcolor[rgb]{0.75,0.0,0.0}{↓}} \scriptsize{2.4}  \cellcolor[HTML]{E6F2FB} 
        & $23.9_{\pm0.17}$ \textbf{\textcolor[rgb]{0.75,0.0,0.0}{↓}} \scriptsize{0.3}  \cellcolor[HTML]{E6F2FB} 
        & $48.6_{\pm0.36}$ \textbf{\textcolor[rgb]{0.75,0.0,0.0}{↓}} \scriptsize{2.4}  \cellcolor[HTML]{E6F2FB} 
        & $41.9_{\pm0.30}$ \textbf{\textcolor[rgb]{0.75,0.0,0.0}{↓}} \scriptsize{3.0}  \cellcolor[HTML]{E6F2FB} 
        & $57.5_{\pm0.67}$ \textbf{\textcolor[rgb]{0.75,0.0,0.0}{↓}} \scriptsize{2.1}  \cellcolor[HTML]{E6F2FB} 
        & $11.3_{\pm0.34}$ \textbf{\textcolor[rgb]{0.75,0.0,0.0}{↓}} \scriptsize{0.6}  \cellcolor[HTML]{E6F2FB} \\
        & {Random Mask \scriptsize{w/ Best FFT}} 
        & $74.9_{\pm0.37}$ 
        & $22.3_{\pm0.23}$ 
        & $48.2_{\pm0.48}$ 
        & $42.7_{\pm0.41}$ 
        & $58.8_{\pm0.96}$ 
        & $11.3_{\pm0.48}$ \\
        & {L1 Mask \scriptsize{w/ Best FFT}} 
        & 75.1 
        & 22.7 
        & 49.1 
        & 44.1 
        & 59.8 
        & 11.5 \\
        & {MFT \scriptsize{w/ Best FFT}} \textbf{(Ours)} \cellcolor[HTML]{FCE7E4} 
        & $\textbf{77.3}_{\pm0.14}$ \textbf{\textcolor[rgb]{0.0, 0.75, 0.0}{↑}} \scriptsize{0.5} \cellcolor[HTML]{FCE7E4} 
        & $\textbf{24.8}_{\pm0.21}$ \textbf{\textcolor[rgb]{0.0, 0.75, 0.0}{↑}} \scriptsize{0.6} \cellcolor[HTML]{FCE7E4} 
        & $\textbf{53.5}_{\pm0.41}$ \textbf{\textcolor[rgb]{0.0, 0.75, 0.0}{↑}} \scriptsize{2.5} \cellcolor[HTML]{FCE7E4} 
        & $\textbf{46.8}_{\pm0.40}$ \textbf{\textcolor[rgb]{0.0, 0.75, 0.0}{↑}} \scriptsize{1.9} \cellcolor[HTML]{FCE7E4} 
        & $\textbf{65.6}_{\pm0.70}$ \textbf{\textcolor[rgb]{0.0, 0.75, 0.0}{↑}} \scriptsize{6.0} \cellcolor[HTML]{FCE7E4} 
        & $\textbf{13.7}_{\pm0.32}$ \textbf{\textcolor[rgb]{0.0, 0.75, 0.0}{↑}} \scriptsize{1.8} \cellcolor[HTML]{FCE7E4} \\
        \hline
        \multirow{8}{*}{\rotatebox[origin=c]{90}{\textbf{Qwen3-1.7B}}}
        & \textbf{Pre-Trained Model}
        & 58.8 
        & 30.2 
        & 38.4 
        & 35.4 
        & 30.3 
        & 9.4 \\
        & Best LoRA \cellcolor[HTML]{EAFBF0} 
        & $63.5_{\pm0.32}$ \cellcolor[HTML]{EAFBF0} 
        & $33.8_{\pm0.25}$ \cellcolor[HTML]{EAFBF0} 
        & $44.5_{\pm0.38}$ \cellcolor[HTML]{EAFBF0} 
        & $39.6_{\pm0.49}$ \cellcolor[HTML]{EAFBF0} 
        & $48.2_{\pm0.75}$ \cellcolor[HTML]{EAFBF0} 
        & $10.6_{\pm0.58}$ \cellcolor[HTML]{EAFBF0} \\
        & Continued LoRA \cellcolor[HTML]{EAFBF0} 
        & $61.8_{\pm0.40}$ \cellcolor[HTML]{EAFBF0} 
        & $32.4_{\pm0.22}$ \cellcolor[HTML]{EAFBF0} 
        & $42.1_{\pm0.52}$ \cellcolor[HTML]{EAFBF0} 
        & $38.0_{\pm0.43}$ \cellcolor[HTML]{EAFBF0} 
        & $44.7_{\pm0.69}$ \cellcolor[HTML]{EAFBF0} 
        & $9.8_{\pm0.33}$ \cellcolor[HTML]{EAFBF0} \\
        & \textbf{Best FFT} \cellcolor[HTML]{E6F2FB} 
        & $68.7_{\pm0.21}$ \cellcolor[HTML]{E6F2FB} 
        & $36.5_{\pm0.20}$ \cellcolor[HTML]{E6F2FB} 
        & $50.5_{\pm0.44}$ \cellcolor[HTML]{E6F2FB} 
        & $44.1_{\pm0.35}$ \cellcolor[HTML]{E6F2FB} 
        & $55.6_{\pm0.78}$ \cellcolor[HTML]{E6F2FB} 
        & $11.4_{\pm0.41}$ \cellcolor[HTML]{E6F2FB} \\
        & {Continued FFT} \cellcolor[HTML]{E6F2FB} 
        & $66.5_{\pm0.19}$ \textbf{\textcolor[rgb]{0.75,0.0,0.0}{↓}} \scriptsize{2.1} \cellcolor[HTML]{E6F2FB}
        & $35.2_{\pm0.19}$ \textbf{\textcolor[rgb]{0.75,0.0,0.0}{↓}} \scriptsize{1.3} \cellcolor[HTML]{E6F2FB}
        & $47.4_{\pm0.41}$ \textbf{\textcolor[rgb]{0.75,0.0,0.0}{↓}} \scriptsize{2.6} \cellcolor[HTML]{E6F2FB}
        & $42.0_{\pm0.34}$ \textbf{\textcolor[rgb]{0.75,0.0,0.0}{↓}} \scriptsize{2.1} \cellcolor[HTML]{E6F2FB}
        & $52.1_{\pm0.76}$ \textbf{\textcolor[rgb]{0.75,0.0,0.0}{↓}} \scriptsize{3.5} \cellcolor[HTML]{E6F2FB}
        & $10.6_{\pm0.39}$ \textbf{\textcolor[rgb]{0.75,0.0,0.0}{↓}} \scriptsize{0.8} \cellcolor[HTML]{E6F2FB}\\
        & {Random Mask \scriptsize{w/ Best FFT}} 
        & $61.5_{\pm0.42}$ 
        & $34.1_{\pm0.27}$ 
        & $47.6_{\pm0.54}$ 
        & $42.0_{\pm0.46}$ 
        & $53.0_{\pm0.80}$ 
        & $10.5_{\pm0.55}$ \\
        & {L1 Mask \scriptsize{w/ Best FFT}} 
        & 66.2 
        & 35.3 
        & 48.5 
        & 43.0 
        & 54.0 
        & 10.8 \\
        & {MFT \scriptsize{w/ Best FFT}} \textbf{(Ours)} \cellcolor[HTML]{FCE7E4}
        & $\textbf{70.8}_{\pm0.17}$ \textbf{\textcolor[rgb]{0.0, 0.75, 0.0}{↑}} \scriptsize{2.1}  \cellcolor[HTML]{FCE7E4}
        & $\textbf{38.3}_{\pm0.24}$ \textbf{\textcolor[rgb]{0.0, 0.75, 0.0}{↑}} \scriptsize{1.8}  \cellcolor[HTML]{FCE7E4}
        & $\textbf{52.0}_{\pm0.46}$ \textbf{\textcolor[rgb]{0.0, 0.75, 0.0}{↑}} \scriptsize{1.5}  \cellcolor[HTML]{FCE7E4}
        & $\textbf{46.3}_{\pm0.45}$ \textbf{\textcolor[rgb]{0.0, 0.75, 0.0}{↑}} \scriptsize{2.2}  \cellcolor[HTML]{FCE7E4}
        & $\textbf{59.7}_{\pm0.80}$ \textbf{\textcolor[rgb]{0.0, 0.75, 0.0}{↑}} \scriptsize{4.1}  \cellcolor[HTML]{FCE7E4}
        & $\textbf{12.4}_{\pm0.36}$ \textbf{\textcolor[rgb]{0.0, 0.75, 0.0}{↑}} \scriptsize{1.0}  \cellcolor[HTML]{FCE7E4} \\
        \hline
        \multirow{8}{*}{\rotatebox[origin=c]{90}{\textbf{Qwen3-14B}}} 
        & \textbf{Pre-Trained Model}
        & 81.0 
        & 45.1 
        & 56.1 
        & 52.4 
        & 45.2 
        & 29.5 \\
        & Best LoRA \cellcolor[HTML]{EAFBF0} 
        & $83.5_{\pm0.24}$ \cellcolor[HTML]{EAFBF0} 
        & $46.7_{\pm0.19}$ \cellcolor[HTML]{EAFBF0} 
        & $66.5_{\pm0.29}$ \cellcolor[HTML]{EAFBF0} 
        & $55.8_{\pm0.40}$ \cellcolor[HTML]{EAFBF0} 
        & $57.5_{\pm0.82}$ \cellcolor[HTML]{EAFBF0} 
        & $32.0_{\pm0.47}$ \cellcolor[HTML]{EAFBF0} \\
        & Continued LoRA \cellcolor[HTML]{EAFBF0} 
        & $82.2_{\pm0.30}$ \cellcolor[HTML]{EAFBF0} 
        & $45.5_{\pm0.17}$ \cellcolor[HTML]{EAFBF0} 
        & $63.3_{\pm0.42}$ \cellcolor[HTML]{EAFBF0} 
        & $53.7_{\pm0.36}$ \cellcolor[HTML]{EAFBF0} 
        & $53.1_{\pm0.79}$ \cellcolor[HTML]{EAFBF0} 
        & $30.4_{\pm0.27}$ \cellcolor[HTML]{EAFBF0} \\
        & \textbf{Best FFT} \cellcolor[HTML]{E6F2FB} 
        & $85.9_{\pm0.16}$ \cellcolor[HTML]{E6F2FB} 
        & $48.9_{\pm0.16}$ \cellcolor[HTML]{E6F2FB} 
        & $75.0_{\pm0.35}$ \cellcolor[HTML]{E6F2FB} 
        & $61.6_{\pm0.28}$ \cellcolor[HTML]{E6F2FB} 
        & $70.3_{\pm0.62}$ \cellcolor[HTML]{E6F2FB} 
        & $39.6_{\pm0.33}$ \cellcolor[HTML]{E6F2FB} \\
        & {Continued FFT} \cellcolor[HTML]{E6F2FB} 
        & $83.6_{\pm0.15}$ \textbf{\textcolor[rgb]{0.75,0.0,0.0}{↓}} \scriptsize{2.2} \cellcolor[HTML]{E6F2FB}
        & $47.6_{\pm0.15}$ \textbf{\textcolor[rgb]{0.75,0.0,0.0}{↓}} \scriptsize{1.3} \cellcolor[HTML]{E6F2FB}
        & $70.8_{\pm0.32}$ \textbf{\textcolor[rgb]{0.75,0.0,0.0}{↓}} \scriptsize{3.6} \cellcolor[HTML]{E6F2FB}
        & $58.0_{\pm0.27}$ \textbf{\textcolor[rgb]{0.75,0.0,0.0}{↓}} \scriptsize{3.6} \cellcolor[HTML]{E6F2FB}
        & $68.5_{\pm0.60}$ \textbf{\textcolor[rgb]{0.75,0.0,0.0}{↓}} \scriptsize{1.8} \cellcolor[HTML]{E6F2FB}
        & $38.8_{\pm0.31}$ \textbf{\textcolor[rgb]{0.75,0.0,0.0}{↓}} \scriptsize{0.8} \cellcolor[HTML]{E6F2FB}\\
        & {Random Mask \scriptsize{w/ Best FFT}} 
        & $83.8_{\pm0.33}$ 
        & $47.5_{\pm0.20}$ 
        & $71.5_{\pm0.43}$ 
        & $59.0_{\pm0.37}$ 
        & $64.5_{\pm0.86}$ 
        & $34.8_{\pm0.43}$ \\
        & {L1 Mask \scriptsize{w/ Best FFT}} 
        & 84.2 
        & 48.0 
        & 71.9 
        & 59.5 
        & 65.0 
        & 35.2 \\
        & {MFT \scriptsize{w/ Best FFT}} \textbf{(Ours)} \cellcolor[HTML]{FCE7E4}
        & $\textbf{86.9}_{\pm0.12}$ \textbf{\textcolor[rgb]{0.0, 0.75, 0.0}{↑}} \scriptsize{1.0}  \cellcolor[HTML]{FCE7E4}
        & $\textbf{50.3}_{\pm0.18}$ \textbf{\textcolor[rgb]{0.0, 0.75, 0.0}{↑}} \scriptsize{1.4}  \cellcolor[HTML]{FCE7E4}
        & $\textbf{76.0}_{\pm0.37}$ \textbf{\textcolor[rgb]{0.0, 0.75, 0.0}{↑}} \scriptsize{1.0}  \cellcolor[HTML]{FCE7E4}
        & $\textbf{63.7}_{\pm0.36}$ \textbf{\textcolor[rgb]{0.0, 0.75, 0.0}{↑}} \scriptsize{2.1}  \cellcolor[HTML]{FCE7E4}
        & $\textbf{76.8}_{\pm0.63}$ \textbf{\textcolor[rgb]{0.0, 0.75, 0.0}{↑}} \scriptsize{6.5}  \cellcolor[HTML]{FCE7E4}
        & $\textbf{42.1}_{\pm0.29}$ \textbf{\textcolor[rgb]{0.0, 0.75, 0.0}{↑}} \scriptsize{2.5}  \cellcolor[HTML]{FCE7E4} \\
        \hline
    \end{tabular}
    }
    \vspace{-4mm}
    \label{tab:exps_main}
\end{table*}

\subsection{Main Results}\label{sec:exps_main_results}
As mentioned in Sec.~\ref{sec:exps_pre}, MFT primarily uses a local masking strategy. Therefore, we begin with a proof-of-concept study to assess whether MFT has potential for further improvements (Fig.~\ref{fig:exps_local_ablation}).
This ablation also helps us identify the most effective components of the well-trained MFT model.
Accordingly, we deploy MFT locally across different configurations and compare it with strong fine-tuning baselines.

\noindent \textbf{Proof-of-Concept Study of Local MFT.}
This series of ablations aims to assess whether the MFT can improve the model's capability after FFT.
If so, we expect a comprehensive ablation to also detect the sensitivity of different model components to MFT.
Specifically, we start from the LLaMA and Qwen backbones after domain-specific FFT and perform MFT on a small subset of each domain.
Specifically, for each domain, we use 10\% of its corresponding fine-tuning set.
To swap the model, we group 4 or 8 consecutive layers to ablate from shallow to deep layers.
For each group, we use MFT to learn a local mask while keeping the other layers fixed to assess model performance.
For example, for layers 4-7, this is formally achieved by setting $\Theta = \{\theta_l\}$, where $l \in \{4,5,6,7\}$.
Within each group, we set the masking ratio to 10\% to learn the mask, meaning we aim to learn and remove 10\% of the parameters while keeping the remaining 90\%.

Fig.~\ref{fig:exps_local_ablation} shows the detailed ablation results.
We find that 1) compared with FFT (black dashed line), the MFT has the potential to further improve for each setting, even if they may have different proper layer-wise groups for MFT;
2) A smaller layer-wise group (4-layer) is better for MFT compared with a larger one (8-layer). However, different model families have different trends. For the LLaMA family, shallow (0-7) and mid-to-rear (20-27) partitions often respond more positively to MFT than other partitions, whereas the Qwen family shows a more consistent trend across groups.
Such observation confirms the MFT's potential to further improve models after FFT and indicates promising partitions for mask learning.
Concretely, for LLaMA2-7B, we use 4-7, 20-23, and 0-3 for math, coding, and instruction domains, respectively, for MFT.
For LLaMA3.1-8B, we use 24-27, 20-23, and 24-27 for the three domains, respectively.
For Qwen3-1.7B, we use 24-27, 12-15, and 4-7 for the three domains, respectively.
For Qwen3-14B, we use 37-40, 32-35, and 7-10 for the three domains, respectively.
For all full-dataset MFT results reported in the rest of this paper, we follow the proof-of-concept study and use a constant 10\% sparse ratio.

Notably, 1) Since this ablation is only for quick intuition and a sanity check of MFT effectiveness, we only use a 10\% dataset subset here.
According to these insights, MFT is trained using complete fine-tuning datasets for more performance gain;
2) Since larger layer groups (e.g., 16) achieve a significant performance drop, we omit their results for a better illustration and leave them in the Appendix.

\begin{figure*}[t]
    \centering
    \includegraphics[width=1\linewidth]{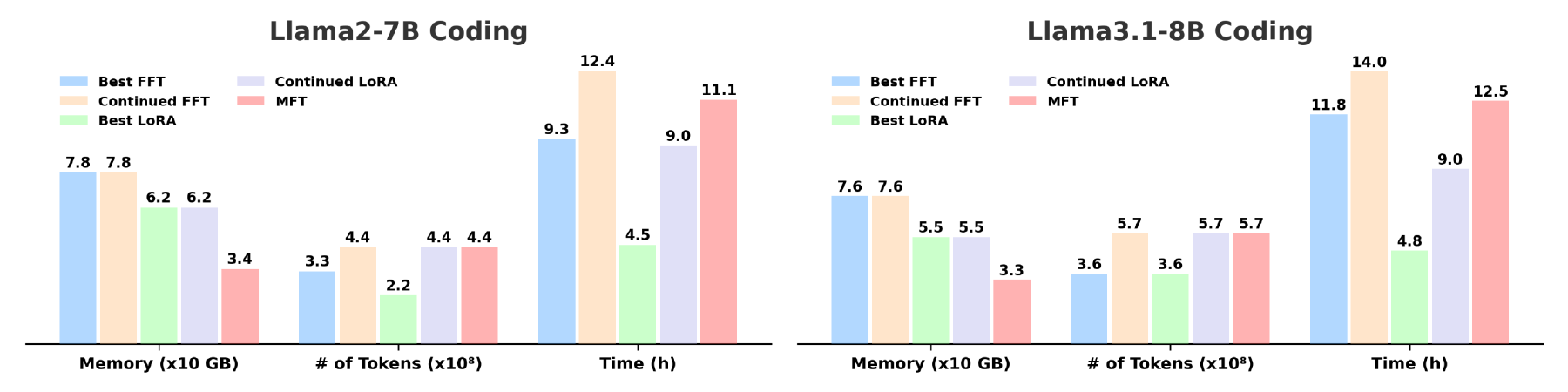}
    \caption{Training cost comparisons of GPU memory, token usage, and training time.
    We use the coding domain as an example to compare MFT with other FFT and LoRA baselines.
    }
    \label{fig:exps_efficiency}
\end{figure*}

\begin{figure*}[t]
    \centering
    \includegraphics[width=1\linewidth]{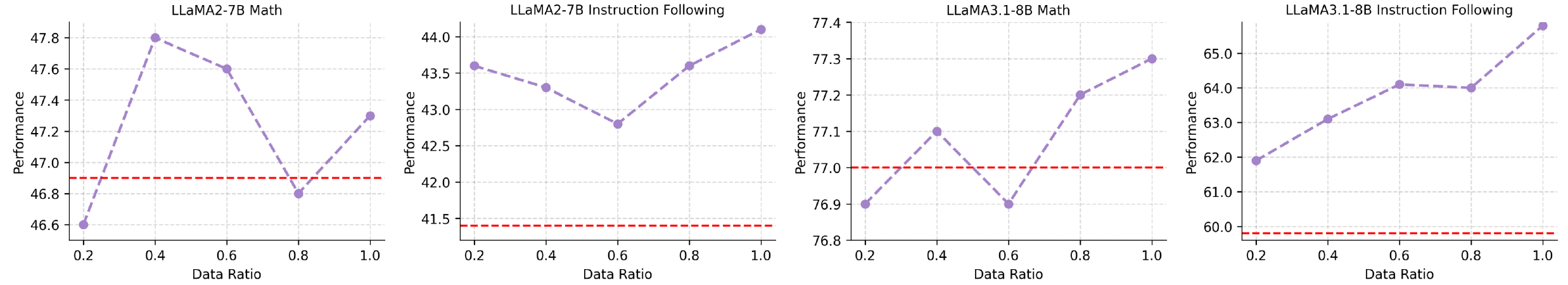}
    \caption{Data ratio ablation.
    We use math and instruction domains on LLaMA2 and LLaMA3.1.
    Compared with Best FFT (red dashed line), MFT (purple) consistently improves performance on the full dataset but may still yield a promising gain with less data.
    }
    \label{fig:exps_data_ratio_ablation}
    \vspace{-4mm}
\end{figure*}

\begin{figure*}[t]
    \centering
    \includegraphics[width=1\linewidth]{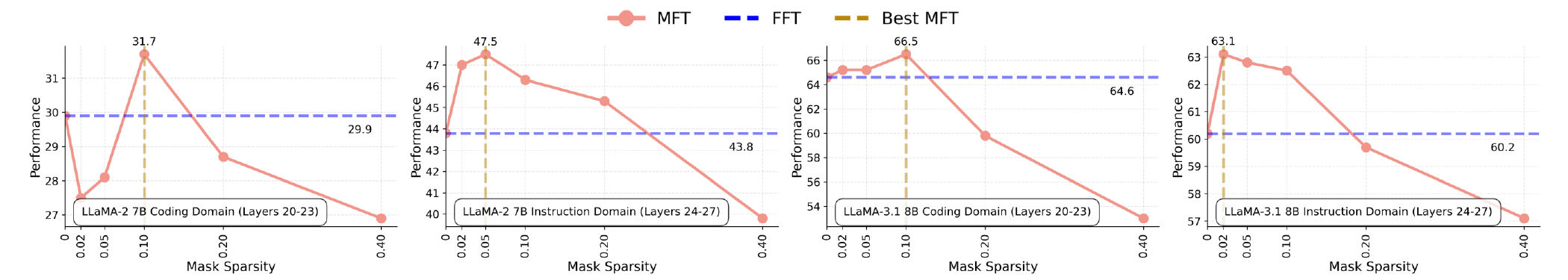}
    \caption{Masking ratio ablation.
    We use coding and instruction domains on LLaMA2 and LLaMA3.1.
    We observe that the original 10\% ratio works well for coding but not for instruction, indicating that the masking ratio matters and that MFT has more potential.
    }
    \label{fig:exps_mask_ratio_ablation}
\end{figure*}

\begin{figure*}[t]
    \centering
    \includegraphics[width=1\linewidth]{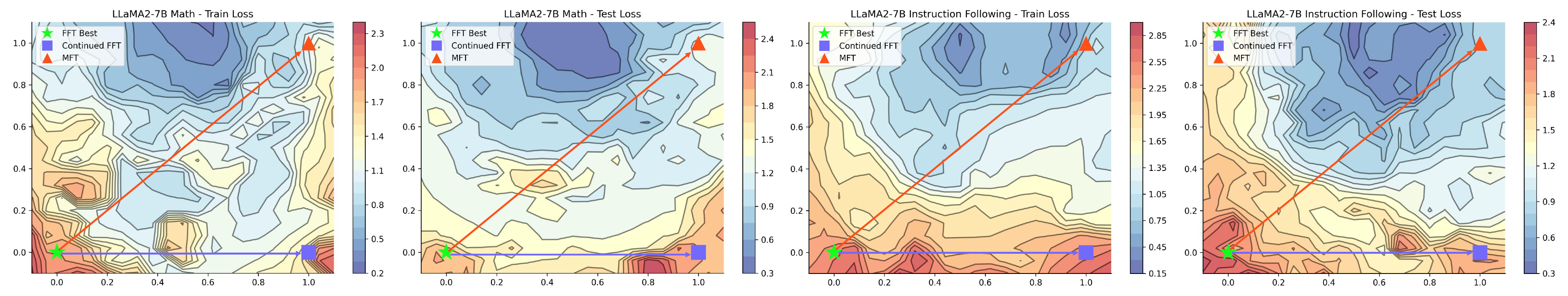}
    \caption{The loss landscape visualization on math and instruction-following domains using LLaMA2-7B.
    Such visualization indicates that the proposed MFT further refines the Best FFT model, improving optimization and generalization.
    }
    \vspace{-4mm}
    \label{fig:loss_surface}
\end{figure*}

\noindent \textbf{Comparisons.}
Tab.~\ref{tab:exps_main} presents our main comparisons using the LLaMA and Qwen backbones.
We start with a pre-trained model, then apply two FFT settings, using the fine-tuning dataset for both specific and mixed domains for each backbone.
For each setting, we perform sufficient FFTs and report the best and continued versions (in the blue block). We do the same for LoRA (green block).
We also include random and L1 masks as two vanilla masking baselines (white block).
MFT (red block) starts from the best-performing FFT and further improves performance.
All three masking methods are based on the best FFT as a strong starting point.
Among them, two critical baselines are the best FFT and continued FFT: the former serves as a competitive baseline, and the latter as a sanity check for potential performance drops due to overfitting, thereby demonstrating MFT's effectiveness.
We train our MFT only using domain-specific datasets for both FFT scenarios (domain-specific and mixed-up).
Additionally, we compare the training cost in Fig.~\ref{fig:exps_efficiency} across GPU memory usage, token count, and overall training time on A100 GPUs.
We use the coding domain as an example. Additional comparisons are provided in the Appendix.
Compared with the most important baseline, Best FFT, MFT adds very little overhead in token usage and training time.
Compared with the Continued FFT, which incurs at least as much, if not more, training overhead, the Best FFT and MFT incur lower training costs while achieving performance improvements.
Since MFT fixes all model parameters and performs only local mask learning, it always has an advantage in memory usage.

Accordingly, we conclude:
1) Since the Continued FFT causes a performance drop and the Best FFT outperforms other comparison methods, it is reasonable to set it as our strong baseline.
2) MFT achieves promising and consistent improvements across different backbones, domains, and FFT configurations.
Accordingly, it upgrades the current routine to a new LLM fine-tuning protocol by learning and applying a mask on a well-trained model.
Moreover, the other two masking baselines generally degrade performance, particularly compared with the consistent improvements achieved by MFT, indicating that MFT is not trivial and warrants further exploration.

\subsection{Analyses}\label{sec:exps_ablation}
\noindent \textbf{Theoretical Analysis.}
We follow PAC-Bayes theory for a generalization upper bound~\citep{mcallester1998some}, which has also been widely explored for neural networks~\citep{arora2018stronger}, to provide information-theoretic intuition.
We focus on the performance gain from the model after the Best FFT to the model after MFT.
Since our MFT uses the same training objective and dataset as FFT, this comparison is fair and yields a purely model-generalization upper bound.

Given $n$ as the size of the training set, $\delta$ as the degree of confidence, for any hypothesis $h$, which represents the model here, we have the PAC-Bayes upper bound given by
\begin{equation}\label{eq:pac-bayes}
L(h) \leq L_S(h) + \Phi(C(h)), \;\;
\Phi(u) = \sqrt{\frac{u + \ln \tfrac{1}{\delta}}{2(n-1)}}.
\end{equation}
$L(h)$ and $L_S(h)$ represent training and test loss, respectively. 
$\Phi(C(h))$ is an additional complexity term to describe the model code length, 
where $C(\cdot)$ represents the encoding process.
Then, we have
\begin{equation}\label{eq:Lfft}
\begin{aligned}
&L(h_{\mathrm{FFT}}) \leq L_S(h_{\mathrm{FFT}}) + \Phi(C_{\mathrm{FFT}}) \\
&\;\Rightarrow U(h_{\mathrm{FFT}}) = L_S(h_{\mathrm{FFT}}) + \Phi(C_{\mathrm{FFT}}),
\end{aligned}
\end{equation}
\begin{equation}\label{eq:Lmft}
\begin{aligned}
&L(h_{\mathrm{MFT}}) \leq L_S(h_{\mathrm{MFT}}) + \Phi(C_{\mathrm{MFT}}) \\
&\;\Rightarrow U(h_{\mathrm{MFT}}) = L_S(h_{\mathrm{MFT}}) + \Phi(C_{\mathrm{MFT}}),
\end{aligned}
\end{equation}
where $U(\cdot)$ means the loss upper bound of $h$. If we take the difference on both sides, we have
\begin{equation}\label{eq:diff}
\begin{aligned}
&U(h_{\mathrm{MFT}}) - U(h_{\mathrm{FFT}}) \\
&\quad = \big[ L_S(h_{\mathrm{MFT}}) - L_S(h_{\mathrm{FFT}}) \big] \\
&\quad + \big[ \Phi(C_{\mathrm{MFT}}) - \Phi(C_{\mathrm{FFT}}) \big],
\end{aligned}
\end{equation}
revised as
\begin{equation}\label{eq:revised}
U(h_{\mathrm{MFT}}) - U(h_{\mathrm{FFT}})
= \Delta_{\mathrm{train}} + \Delta_{\mathrm{complexity}}.
\end{equation}
We numerically show $\Delta_{\mathrm{train}} + \Delta_{\mathrm{complexity}} < 0$.
Details are provided in the Appendix.

\noindent \textbf{Data Ratio Ablation.}
We ablate data ratios for MFT.
Fig.~\ref{fig:exps_data_ratio_ablation} shows performance trends in the math and instruction-following domains on LLaMA2-7B and LLaMA3.1-8B (others are in the supplementary).
We find that MFT (purple), when trained on the full dataset, consistently improves the model compared with the Best FFT (red dashed line). However, less data may also perform competitively and more efficiently.

\noindent \textbf{Masking Ratio Ablation.}
We ablate the masking ratio for local MFT.
We use a 10\% masking ratio across all settings (Tab.~\ref{tab:exps_main}), which may not be optimal.
Here, we define a series of masking ratios for the ablation study.
Specifically, we use all three domains on two backbones and set the masking ratio from 0.02 to 0.4.
Fig.~\ref{fig:exps_mask_ratio_ablation} shows results for coding and instruction, and the math results are in the supplementary material.
We find that our original 10\% ratio works well for certain coding cases.
However, it is not optimal for the instruction domain, where a smaller ratio works better than 10\%.
This suggests that the MFT masking ratio affects final performance.

\noindent \textbf{Mask Fine-Tuning Loss Landscape.}
Loss surface visualization is commonly used to analyze the model optimization process and provide insights into the model's status from a loss perspective.
Thus, we visualize the MFT loss surface to illustrate its training dynamics.
Fig.~\ref{fig:loss_surface} shows the surface of LLaMA2-7B on math and instruction-following domains.
The Best FFT, continued FFT, and MFT are represented by the green star, the purple rectangle, and the red triangle, respectively.
We find that MFT always optimizes the model to a better status than the Best FFT start point and its continued version in both training and test scenarios.
We provide additional theoretical analysis in the Appendix.

\section{Related Work}\label{sec:literature}
This work explores optimizing a well-trained large language model (LLM) by learning a binary mask and proposes mask fine-tuning (MFT), an upgrade to the current LLM fine-tuning protocol.
We summarize the relevant literature on our work from (1) LLM pretraining and fine-tuning and (2) sparse network perspectives, emphasizing their correlations and differences.

\subsection{LLMs Pre-training \& Fine-tuning}
LLM pre-training leads to significant performance gains in general language model capacity for both understanding~\citep{devlin2018bert, raffel2020exploring, liu2019roberta, he2020deberta} and generation~\citep{radford2019language, brown2020language, achiam2023gpt, dubey2024llama, touvron2023llama}.
Massive language corpora with large-scale pretraining enable them to serve as strong language backbones with basic common knowledge.
Following pretraining, downstream fine-tuning on small but high-quality target datasets further enhances pre-trained models with the specific capacities of target domains~\citep{hendrycks2021measuring, yu2023metamath, cobbe2021training, chen2021evaluating, austin2021program, zhou2023instruction, wang2023clinicalgpt, nguyen2023brief}.
Different fine-tuning optimization strategies deliver various LLM properties, such as supervised fine-tuning (SFT), which follows a pre-training objective to involve extra domain knowledge into LLM~\citep{taori2023alpaca, chiang2023vicuna, wang2022self}, and policy-based fine-tuning methods better align the human preference with LLM~\citep{schulman2017proximal, rafailov2024direct, ivison2024unpacking, ouyang2022training, dai2023safe}.
We propose Mask Fine-Tuning (MFT) as a post-fine-tuning strategy to further refine the fine-tuned LLM, especially for SFT in this study.
This strategy can be flexibly deployed after any fine-tuned model to propose a new protocol for improving LLMs.

\subsection{Sparse Networks}
Model sparsity in neural networks is close to neural network pruning~\citep{liu2018rethinking, molchanov2016pruning,wang2021recent} for model compression~\citep{han2015deep, iandola2016squeezenet} and acceleration~\citep{han2016eie,wang2020neural,wang2022trainability,ma2023llm, fang2023structural}.
Our work overlaps with network pruning, as both aim to identify a model partition with a given sparsity. However, pruning aims to remove a large portion of the model for efficiency, whereas ours focuses on removing parameters that are irrelevant or even harmful to the model's capacity, potentially requiring only a small subset of the model.
More importantly, pruning methods inevitably entail a performance loss in the trained model, whereas ours focuses on further enhancing it.
In addition, the sparsity concept is also involved in other machine learning fields such as network optimization~\citep{srivastava2014dropout,srinivas2017training,baldi2013understanding}, model architecture design~\citep{shazeer2017outrageously, zoph2016neural,elsken2019neural}, lottery ticket hypothesis~\citep{frankle2018lottery,zhou2019deconstructing,you2019drawing,bai2022dual}, neural network mechanism~\citep{wortsman2020supermasks, ramanujan2020s,bai2022parameter}, etc.
In our study, we introduce sparsity to investigate whether all parameters are beneficial for a well-trained LLM and to improve the model by learning and applying a mask, proposing a new approach to LLM fine-tuning.

\section{Conclusion}\label{sec:conclusion}

In this paper, we challenge the necessity of structural integrity in large language models (LLMs) by examining whether a well-trained LLM can be further improved by removing certain model parameters.
We propose mask fine-tuning (MFT) to explore this using a model with sufficient full fine-tuning (FFT), which consistently yields further improvements, generally enhances LLM capabilities, and naturally upgrades the LLM fine-tuning pipeline.
Comprehensive experiments with detailed ablations support our conclusion, drawing on exploration intuitions, covering different pre-trained LLM backbones, downstream domains, and fine-tuning configurations.
Meanwhile, MFT extends model sparsity beyond compression, enabling it in a broader range of scenarios and inspiring further work.

\newpage







\bibliography{iclr2027_conference}
\bibliographystyle{iclr2027_conference}

\newpage

\appendix
\section*{Appendix}

\begin{table*}[htbp]
    \renewcommand{\arraystretch}{1.25}
    \caption{The information on the training and test datasets used in our experiments.}
    \centering
    \resizebox{\textwidth}{!}{
    \begin{tabular}{c|c|c|c}
        \hline
        \textbf{Domain} & \textbf{Train / Test} & \textbf{Dataset Name} & \textbf{$\#$ of Samples} \\
        \hline
        \multirow{7}{*}{Math} 
          & \multirow{5}{*}{Train} & Tulu 3 Persona MATH~\citep{lambert2025tulu3pushingfrontiers} & 149,960 \\
        \cline{3-4}
          &                      & Tulu 3 Persona GSM~\citep{lambert2025tulu3pushingfrontiers}  & 49,980 \\
        \cline{3-4}
          &                      & Tulu 3 Persona Algebra~\citep{lambert2025tulu3pushingfrontiers} & 20,000 \\
        \cline{3-4}
          &                      & MetaMathQA~\citep{yu2023metamath} & 395,000 \\
        \cline{3-4}
          &                      & NuminaMath-TIR~\citep{numina_math_datasets} & 64,312 \\
        \cline{2-4}
          & \multirow{2}{*}{Test} & GSM8K~\citep{cobbe2021training} & 1,320 \\
        \cline{3-4}
          &                      & MATH~\citep{2206.14858} & 5,003 \\
        \hline
        \multirow{5}{*}{Coding} 
          & \multirow{3}{*}{Train} & Evol CodeAlpaca~\citep{luo2023wizardcoder} & 107,276 \\
        \cline{3-4}
          &                      & Code-Alpaca~\citep{codealpaca} & 20,022 \\
        \cline{3-4}
          &                      & Tulu 3 Persona Python~\citep{lambert2025tulu3pushingfrontiers} & 34,999 \\
        \cline{2-4}
          & \multirow{2}{*}{Test} & HumanEval~\citep{chen2021evaluating} & 164 \\
        \cline{3-4}
          &                      & HumanEval+~\citep{evalplus} & 164 \\
        \hline
        \multirow{5}{*}{Instruction Following} 
          & \multirow{3}{*}{Train} & OpenAssistant Guanaco~\citep{köpf2023openassistantconversationsdemocratizing} & 7,132 \\
        \cline{3-4}
          &                      & Tulu 3 Persona IF~\citep{lambert2025tulu3pushingfrontiers} & 29,980 \\
        \cline{3-4}
          &                      & Open-Orca~\citep{mukherjee2023orcaprogressivelearningcomplex} & 30,000 \\
        \cline{2-4}
          & \multirow{2}{*}{Test} & IFEval~\citep{zhou2023instruction} & 541 \\
        \cline{3-4}
          &                      & AlpacaEval 2.0~\citep{alpaca_eval} & 805 \\
        \hline
    \end{tabular}
    }
    \label{tab:dataset_info}
\end{table*}

\section{Implementation Details}\label{sec:supp_implementation}
We conduct our experiments on 8*A100 GPUs.
To ensure that full fine-tuning (FFT) is sufficient, we set the maximum number of FFT epochs to 4, as we observe overfitting and a drop in performance.
We fine-tune MFT for 2 additional epochs, starting from the best FFT checkpoint (a total of 4 epochs).
We set the batch size to 8, the accumulation steps to 16, and the weight decay to 0.0 for all settings.
We use learning rates of 2e-5 and 1e-5 for LLaMA and Qwen, with a linear schedule.
The warm-up learning rate starts at 0 and increases linearly, with a warm-up ratio of 0.03 of the total training.
We summarize the training and test datasets used in our experiments in Tab.~\ref{tab:dataset_info}.

\begin{table*}[ht]
    \centering
    \renewcommand{\arraystretch}{1.25}
    \caption{Cross-domain evaluations on LLaMA2-7B under domain-specific FFT setting.
    }
    \setlength{\tabcolsep}{12pt}
    \resizebox{\textwidth}{!}{
    \begin{tabular}{l|c|c|c|c|c|c}
        \hline
        \multirow{2}{*}{\textbf{Method}} & \multicolumn{2}{c|}{\textbf{Math}} & \multicolumn{2}{c|}{\textbf{Coding}} & \multicolumn{2}{c}{\textbf{Instruction Following}} \\
        \cline{2-7}
        & GSM8K & Math & HumanEval & HumanEval+ & IF-Eval & Alpaca-Eval \\
        \hline
        \textbf{Pre-trained Model} & 15.2 & 2.5 & 25.8 & 22.4 & 34.3 & 0.5 \\
        \textbf{Best FFT} [Math] & 46.9 & 6.7 & 17.7 & 15.9 & 29.3 & 1.5 \\
        \textbf{Best FFT} [Coding] & 14.6 & 2.8 & 29.3 & 25.0 & 8.3 & 1.4 \\
        \textbf{Best FFT} [IF] & 25.0 & 2.4 & 16.5 & 13.4 & 41.4 & 1.7 \\
        \hline
        {Continued FFT [Math]} & 45.2 & 5.7 & 19.5 & 17.1 & 33.2 & 1.5 \\
        {Continued FFT [Coding]} & 11.1 & 2.5 & 28.0 & 23.8 & 13.1 & 2.7 \\
        {Continued FFT [IF]} & 23.1 & 2.1 & 13.1 & 12.8 & 37.4 & 2.0 \\
        {Random Mask \scriptsize{w/ Best FFT} [Math]} & 38.4 & 4.9 & 14.8 & 13.1 & 25.5 & 1.0 \\
        {Random Mask \scriptsize{w/ Best FFT} [Coding]} & 12.4 & 2.5 & 28.0 & 24.0 & 7.5 & 1.1 \\
        {Random Mask \scriptsize{w/ Best FFT} [IF]} & 23.0 & 2.3 & 14.4 & 12.2 & 39.4 & 1.5 \\
        {L1 Mask \scriptsize{w/ Best FFT} [Math]} & 42.6 & 5.7 & 16.4 & 14.6 & 28.3 & 1.1 \\
        {L1 Mask \scriptsize{w/ Best FFT} [Coding]} & 12.8 & 2.6 & 28.9 & 24.5 & 7.8 & 1.0 \\
        {L1 Mask \scriptsize{w/ Best FFT} [IF]} & 23.9 & 2.4 & 14.9 & 12.7 & 40.9 & 1.5 \\
        \hline
        \textbf{[Ours]} {Best FFT + Mask [Math]} & \textbf{47.3} & \textbf{7.6} & 16.5 & 14.6 & 30.1 & 0.5 \\
        \textbf{[Ours]} {Best FFT + Mask [Coding]} & 13.8 & 2.8 & \textbf{31.7} & \textbf{28.0} & 8.5 & 2.1 \\
        \textbf{[Ours]} {Best FFT + Mask [IF]} & 23.5 & 2.8 & 15.2 & 11.6 & \textbf{44.1} & \textbf{2.9} \\
        \hline
    \end{tabular}
    }
    \label{tab:supp_exp_llama2}
\end{table*}

\begin{table*}[ht]
    \centering
    \renewcommand{\arraystretch}{1.25}
    \caption{Cross-domain evaluations on LLaMA3.1-8B under domain-specific FFT setting.
    }
    \setlength{\tabcolsep}{12pt}
    \resizebox{\textwidth}{!}{
    \begin{tabular}{l|c|c|c|c|c|c}
        \hline
        \multirow{2}{*}{\textbf{Method}} & \multicolumn{2}{c|}{\textbf{Math}} & \multicolumn{2}{c|}{\textbf{Coding}} & \multicolumn{2}{c}{\textbf{Instruction Following}} \\
        \cline{2-7}
        & GSM8K & Math & HumanEval & HumanEval+ & IF-Eval & Alpaca-Eval \\
        \hline
        \textbf{Pre-trained Model} & 55.9 & 14.6 & 42.1 & 37.8 & 17.6 & 9.0 \\
        \textbf{Best FFT} [Math] & 77.0 & 24.3 & 35.4 & 29.9 & 31.5 & 0.6 \\
        \textbf{Best FFT} [Coding] & 60.7 & 13.8 & 51.2 & 45.1 & 5.3 & 0.9 \\
        \textbf{Best FFT} [IF] & 63.0 & 10.1 & 50.0 & 44.5 & 59.8 & 12.0 \\
        \hline
        {Continued FFT [Math]} & 74.6 & 24.0 & 33.5 & 31.1 & 32.2 & 0.7 \\
        {Continued FFT [Coding]} & 59.6 & 13.1 & 48.8 & 42.1 & 4.8 & 1.4 \\
        {Continued FFT [IF]} & 63.6 & 8.7 & 45.7 & 43.5 & 57.2 & 11.4 \\
        {Random Mask \scriptsize{w/ Best FFT} [Math]} & 74.9 & 22.3 & 34.3 & 28.5 & 30.0 & 0.4 \\
        {Random Mask \scriptsize{w/ Best FFT} [Coding]} & 56.2 & 12.7 & 48.2 & 42.7 & 4.7 & 0.7 \\
        {Random Mask \scriptsize{w/ Best FFT} [IF]} & 61.3 & 9.2 & 45.3 & 41.7 & 58.8 & 11.5 \\
        {L1 Mask \scriptsize{w/ Best FFT} [Math]} & 75.1 & 22.7 & 34.4 & 30.6 & 31.1 & 0.6 \\
        {L1 Mask \scriptsize{w/ Best FFT} [Coding]} & 57.3 & 12.9 & 49.1 & 44.1 & 4.8 & 0.8 \\
        {L1 Mask \scriptsize{w/ Best FFT} [IF]} & 62.3 & 9.3 & 46.1 & 42.4 & 59.8 & 11.3 \\
        \textbf{[Ours]} {Best FFT + Mask [Math]} & \textbf{77.3} & \textbf{24.9} & 36.0 & 32.2 & 31.1 & 0.5 \\
        \textbf{[Ours]} {Best FFT + Mask [Coding]} & 58.6 & 13.6 & \textbf{53.7} & \textbf{47.0} & 4.9 & 0.0 \\
        \textbf{[Ours]} {Best FFT + Mask [Instruction]} & 63.2 & 9.6 & 44.5 & 40.9 & \textbf{65.8} & \textbf{13.8} \\
        \hline
    \end{tabular}
    }
    \label{tab:supp_exp_llama3}
\end{table*}

\section{More Evaluations}\label{sec:supp_main_results}
Tab.~\ref{tab:supp_exp_llama2} and Tab.~\ref{tab:supp_exp_llama3} extend the performance tables in our main draft by adding cross-domain evaluations for our MFT, FFT, and other comparison methods, under domain-specific FFT settings.
The best performance for each setting is highlighted in bold.
These cross-domain results are included to better understand MFT's behavior beyond the target domain.
MFT is designed as a domain-specific post-FFT refinement method: given a well-fine-tuned model and a target domain, it learns a binary mask to identify the subnetwork that is most beneficial for that domain. Therefore, the learned mask may remove weights that are redundant or noisy for the target domain, but those weights may still be useful for other domains. In this sense, cross-domain performance degradation is expected and consistent with MFT's role as a domain-specialized refinement mechanism.


\section{More Ablations and Analyses}\label{sec:supp_ablation}
\noindent \textbf{Multi-domain Settings.}
Tab.~\ref{tab:exps_supp} reports additional results under mixed-domain FFT settings. 
Here, we first obtain the starting checkpoint from a well-tuned mixed-domain FFT model, then apply each post-FFT refinement method, including continued FFT/LoRA and MFT, in the same domain-specific setting for a fair comparison. 
This setting reflects a practical scenario: given a strong mixed-domain fine-tuned checkpoint, can MFT still serve as an effective domain-specific post-FFT refinement method? 
The results show that MFT consistently improves over Best FFT across all six benchmarks on both LLaMA2-7B and LLaMA3.1-8B. 
In contrast, continued FFT usually degrades performance, and random or L1 masking fails to consistently match Best FFT. 
These results suggest that MFT remains effective beyond domain-specific FFT and can refine a mixed-domain model through learned task-aware masks rather than additional optimization steps or heuristic sparsity.

\begin{table*}[t]
    \caption{
    Performance comparison on LLaMA2-7B and LLaMA3.1-8B in multi-domain settings.
    Each part has three blocks containing 1) common LoRA and its continued variant (green), 2) FFT serving as an upper bound and its continued variant (blue), and 3) our MFT (red) with the other two vanilla masking baselines.
    The pre-trained performance serves as a lower bound.
    }
    \centering
    \renewcommand{\arraystretch}{1.25}
    \setlength{\tabcolsep}{12pt}
    \resizebox{\textwidth}{!}{
    \begin{tabular}{l|lllllll}
        \hline
        \multicolumn{2}{c}{\multirow{2}{*}{\textbf{Method}}} 
        & \multicolumn{2}{c}{\textbf{Math}} 
        & \multicolumn{2}{c}{\textbf{Coding}} 
        & \multicolumn{2}{c}{\textbf{Instruction Following}} \\
        \cline{3-8}
        \multicolumn{2}{l}{} 
        & \multicolumn{1}{c}{GSM8K} 
        & \multicolumn{1}{c}{Math} 
        & \multicolumn{1}{c}{HumanEval} 
        & \multicolumn{1}{c}{HumanEval+} 
        & \multicolumn{1}{c}{IF-Eval} 
        & \multicolumn{1}{c}{Alpaca-Eval} \\
        \hline
        \multicolumn{2}{c}{\textbf{Pre-Trained Model}}
        & 15.2 
        & 2.5 
        & 25.8 
        & 22.4 
        & 34.3 
        & 0.5 \\
        \hline
        \multirow{7}{*}{\rotatebox[origin=c]{90}{\textbf{Mixed Domain}}} 
        & Best LoRA \cellcolor[HTML]{EAFBF0} 
        & $40.8_{\pm0.30}$  \cellcolor[HTML]{EAFBF0} 
        & $6.1_{\pm0.28}$  \cellcolor[HTML]{EAFBF0} 
        & $22.6_{\pm0.55}$  \cellcolor[HTML]{EAFBF0} 
        & $18.3_{\pm0.44}$  \cellcolor[HTML]{EAFBF0} 
        & $37.3_{\pm0.81}$  \cellcolor[HTML]{EAFBF0} 
        & $0.7_{\pm0.11}$  \cellcolor[HTML]{EAFBF0} \\
        & Continued LoRA \cellcolor[HTML]{EAFBF0} 
        & $31.9_{\pm0.24}$  \cellcolor[HTML]{EAFBF0} 
        & $4.0_{\pm0.21}$  \cellcolor[HTML]{EAFBF0} 
        & $20.1_{\pm0.42}$  \cellcolor[HTML]{EAFBF0} 
        & $17.1_{\pm0.36}$  \cellcolor[HTML]{EAFBF0} 
        & $31.5_{\pm0.92}$  \cellcolor[HTML]{EAFBF0} 
        & $0.8_{\pm0.09}$  \cellcolor[HTML]{EAFBF0} \\
        & {\textbf{Best FFT}} \cellcolor[HTML]{E6F2FB} 
        & $45.5_{\pm0.18}$ \cellcolor[HTML]{E6F2FB} 
        & $8.1_{\pm0.26}$ \cellcolor[HTML]{E6F2FB} 
        & $29.7_{\pm0.43}$ \cellcolor[HTML]{E6F2FB} 
        & $26.7_{\pm0.38}$ \cellcolor[HTML]{E6F2FB} 
        & $43.6_{\pm0.68}$ \cellcolor[HTML]{E6F2FB} 
        & $1.0_{\pm0.15}$ \cellcolor[HTML]{E6F2FB} \\
        & {Continued FFT} \cellcolor[HTML]{E6F2FB} 
        & $44.1_{\pm0.22}$ \textbf{\textcolor[rgb]{0.75,0.0,0.0}{↓}} \scriptsize{1.4} \cellcolor[HTML]{E6F2FB} 
        & $7.5_{\pm0.27}$ \textbf{\textcolor[rgb]{0.75,0.0,0.0}{↓}} \scriptsize{0.6} \cellcolor[HTML]{E6F2FB} 
        & $21.1_{\pm0.58}$ \textbf{\textcolor[rgb]{0.75,0.0,0.0}{↓}} \scriptsize{8.6} \cellcolor[HTML]{E6F2FB} 
        & $18.1_{\pm0.43}$ \textbf{\textcolor[rgb]{0.75,0.0,0.0}{↓}} \scriptsize{8.6} \cellcolor[HTML]{E6F2FB} 
        & $38.6_{\pm0.74}$ \textbf{\textcolor[rgb]{0.75,0.0,0.0}{↓}} \scriptsize{5.0} \cellcolor[HTML]{E6F2FB} 
        & $1.2_{\pm0.16}$ \textbf{\textcolor[rgb]{0.0, 0.75, 0.0}{↑}}  \scriptsize{0.2} \cellcolor[HTML]{E6F2FB} \\
        & {Random Mask \scriptsize{w/ Best FFT}} 
        & $40.0_{\pm0.33}$ 
        & $6.5_{\pm0.32}$ 
        & $23.7_{\pm0.57}$ 
        & $19.1_{\pm0.49}$ 
        & $30.0_{\pm0.95}$ 
        & $0.9_{\pm0.14}$ \\
        & {L1 Mask \scriptsize{w/ Best FFT}} 
        & 43.3 
        & 7.7 
        & 25.7 
        & 22.3 
        & 32.8 
        & 1.2 \\
        & {MFT \scriptsize{w/ Best FFT}} \textbf{(Ours)} \cellcolor[HTML]{FCE7E4}
        & $\textbf{45.9}_{\pm0.17}$ \textbf{\textcolor[rgb]{0.0, 0.75, 0.0}{↑}} \scriptsize{0.4}  \cellcolor[HTML]{FCE7E4} 
        & $\textbf{8.4}_{\pm0.28}$ \textbf{\textcolor[rgb]{0.0, 0.75, 0.0}{↑}} \scriptsize{0.3}  \cellcolor[HTML]{FCE7E4} 
        & $\textbf{31.5}_{\pm0.46}$ \textbf{\textcolor[rgb]{0.0, 0.75, 0.0}{↑}} \scriptsize{1.8}  \cellcolor[HTML]{FCE7E4} 
        & $\textbf{27.3}_{\pm0.41}$ \textbf{\textcolor[rgb]{0.0, 0.75, 0.0}{↑}} \scriptsize{0.6}  \cellcolor[HTML]{FCE7E4} 
        & $\textbf{46.1}_{\pm0.71}$ \textbf{\textcolor[rgb]{0.0, 0.75, 0.0}{↑}} \scriptsize{2.5}  \cellcolor[HTML]{FCE7E4} 
        & $\textbf{1.9}_{\pm0.18}$ \textbf{\textcolor[rgb]{0.0, 0.75, 0.0}{↑}} \scriptsize{0.9}  \cellcolor[HTML]{FCE7E4} \\
        \hline
        \multicolumn{2}{c}{\textbf{Pre-Trained Model}}
        & 55.9 
        & 14.6 
        & 42.1 
        & 37.8 
        & 17.6 
        & 9.0 \\
        \hline
        \multirow{7}{*}{\rotatebox[origin=c]{90}{\textbf{Multi Domain}}} 
        & Best LoRA \cellcolor[HTML]{EAFBF0} 
        & $72.8_{\pm0.42}$  \cellcolor[HTML]{EAFBF0} 
        & $24.5_{\pm0.29}$  \cellcolor[HTML]{EAFBF0} 
        & $63.1_{\pm0.82}$  \cellcolor[HTML]{EAFBF0} 
        & $55.5_{\pm0.65}$  \cellcolor[HTML]{EAFBF0} 
        & $38.2_{\pm1.18}$  \cellcolor[HTML]{EAFBF0} 
        & $11.0_{\pm0.64}$  \cellcolor[HTML]{EAFBF0} \\
        & Continued LoRA \cellcolor[HTML]{EAFBF0} 
        & $70.4_{\pm0.36}$  \cellcolor[HTML]{EAFBF0} 
        & $24.1_{\pm0.27}$  \cellcolor[HTML]{EAFBF0} 
        & $59.1_{\pm0.75}$  \cellcolor[HTML]{EAFBF0} 
        & $54.3_{\pm0.59}$  \cellcolor[HTML]{EAFBF0} 
        & $32.1_{\pm1.05}$  \cellcolor[HTML]{EAFBF0} 
        & $5.3_{\pm0.32}$  \cellcolor[HTML]{EAFBF0} \\
        & {\textbf{Best FFT}} \cellcolor[HTML]{E6F2FB} 
        & $72.7_{\pm0.16}$  \cellcolor[HTML]{E6F2FB} 
        & $24.4_{\pm0.21}$  \cellcolor[HTML]{E6F2FB} 
        & $64.4_{\pm0.59}$  \cellcolor[HTML]{E6F2FB} 
        & $60.2_{\pm0.48}$  \cellcolor[HTML]{E6F2FB} 
        & $60.0_{\pm0.73}$  \cellcolor[HTML]{E6F2FB} 
        & $11.8_{\pm0.48}$  \cellcolor[HTML]{E6F2FB} \\
        & {Continued FFT}  \cellcolor[HTML]{E6F2FB} 
        & $71.7_{\pm0.14}$ \textbf{\textcolor[rgb]{0.75,0.0,0.0}{↓}} \scriptsize{1.0}  \cellcolor[HTML]{E6F2FB} 
        & $24.2_{\pm0.20}$ \textbf{\textcolor[rgb]{0.75,0.0,0.0}{↓}} \scriptsize{0.2}  \cellcolor[HTML]{E6F2FB} 
        & $62.6_{\pm0.53}$ \textbf{\textcolor[rgb]{0.75,0.0,0.0}{↓}} \scriptsize{1.8}  \cellcolor[HTML]{E6F2FB} 
        & $58.3_{\pm0.44}$ \textbf{\textcolor[rgb]{0.75,0.0,0.0}{↓}} \scriptsize{1.9}  \cellcolor[HTML]{E6F2FB} 
        & $59.7_{\pm0.80}$ \textbf{\textcolor[rgb]{0.75,0.0,0.0}{↓}} \scriptsize{0.3}  \cellcolor[HTML]{E6F2FB} 
        & $8.8_{\pm0.38}$ \textbf{\textcolor[rgb]{0.75,0.0,0.0}{↓}} \scriptsize{3.0}  \cellcolor[HTML]{E6F2FB} \\
        & {Random Mask \scriptsize{w/ Best FFT}} 
        & $72.5_{\pm0.39}$ 
        & $23.2_{\pm0.28}$ 
        & $58.5_{\pm0.73}$ 
        & $55.5_{\pm0.62}$ 
        & $57.7_{\pm1.22}$ 
        & $10.0_{\pm0.51}$ \\
        & {L1 Mask \scriptsize{w/ Best FFT}} 
        & 72.5 
        & 24.1 
        & 63.4 
        & 59.1 
        & 58.9 
        & 10.7 \\
        & {MFT \scriptsize{w/ Best FFT}} \textbf{(Ours)} \cellcolor[HTML]{FCE7E4} 
        & $\textbf{73.6}_{\pm0.20}$ \textbf{\textcolor[rgb]{0.0, 0.75, 0.0}{↑}} \scriptsize{0.9} \cellcolor[HTML]{FCE7E4} 
        & $\textbf{25.2}_{\pm0.23}$ \textbf{\textcolor[rgb]{0.0, 0.75, 0.0}{↑}} \scriptsize{0.8} \cellcolor[HTML]{FCE7E4} 
        & $\textbf{66.3}_{\pm0.51}$ \textbf{\textcolor[rgb]{0.0, 0.75, 0.0}{↑}} \scriptsize{1.9} \cellcolor[HTML]{FCE7E4} 
        & $\textbf{61.4}_{\pm0.50}$ \textbf{\textcolor[rgb]{0.0, 0.75, 0.0}{↑}} \scriptsize{1.2} \cellcolor[HTML]{FCE7E4} 
        & $\textbf{62.3}_{\pm0.69}$ \textbf{\textcolor[rgb]{0.0, 0.75, 0.0}{↑}} \scriptsize{2.3} \cellcolor[HTML]{FCE7E4} 
        & $\textbf{12.2}_{\pm0.34}$ \textbf{\textcolor[rgb]{0.0, 0.75, 0.0}{↑}} \scriptsize{0.4} \cellcolor[HTML]{FCE7E4} \\
        \hline
    \end{tabular}
    }
    \label{tab:exps_supp}
\end{table*}

\begin{wraptable}{r}{0.58\columnwidth}
    \centering
    \vspace{-10pt}
    \footnotesize
    \caption{
    Initial exploration of global MFT on the math domain using LLaMA2-7B and LLaMA3.1-8B backbones.
    We observe better performance on LLaMA2-7B, especially on GSM8K, but lower performance on LLaMA3.1-8B.
    }
    \vspace{2pt}

    \renewcommand{\arraystretch}{1.15}%
    \resizebox{\linewidth}{!}{%
    \begin{tabular}{lcccc}
        \hline
        \multirow{2}{*}{\textbf{Method}} 
        & \multicolumn{2}{c}{\textbf{LLaMA2-7B}} 
        & \multicolumn{2}{c}{\textbf{LLaMA3.1-8B}} \\
        \cline{2-5}
        & GSM8K & Math & GSM8K & Math \\
        \hline

        {Pre-Trained Model} 
        & 15.2 
        & 2.5 
        & 55.9 
        & 14.6 \\
        \hline

        {Best FFT} \cellcolor[HTML]{E6F2FB} 
        & 46.9 \cellcolor[HTML]{E6F2FB} 
        & 6.7 \cellcolor[HTML]{E6F2FB} 
        & 77.0 \cellcolor[HTML]{E6F2FB} 
        & 24.3 \cellcolor[HTML]{E6F2FB} \\

        {Continued FFT} \cellcolor[HTML]{E6F2FB} 
        & 45.2 \textbf{\textcolor[rgb]{0.75,0.0,0.0}{↓}} \scriptsize{1.7} \cellcolor[HTML]{E6F2FB} 
        & 5.7 \textbf{\textcolor[rgb]{0.75,0.0,0.0}{↓}} \scriptsize{1.0} \cellcolor[HTML]{E6F2FB} 
        & 74.6 \textbf{\textcolor[rgb]{0.75,0.0,0.0}{↓}} \scriptsize{2.4} \cellcolor[HTML]{E6F2FB} 
        & 24.0 \textbf{\textcolor[rgb]{0.75,0.0,0.0}{↓}} \scriptsize{0.3} \cellcolor[HTML]{E6F2FB} \\

        {Global MFT \scriptsize{w/ Best FFT}} \textbf{(Ours)} \cellcolor[HTML]{FCE7E4} 
        & 49.0 \textbf{\textcolor[rgb]{0.0,0.75,0.0}{↑}} \scriptsize{2.1} \cellcolor[HTML]{FCE7E4} 
        & 7.1 \textbf{\textcolor[rgb]{0.0,0.75,0.0}{↑}} \scriptsize{0.4} \cellcolor[HTML]{FCE7E4} 
        & 74.1 \textbf{\textcolor[rgb]{0.75,0.0,0.0}{↓}} \scriptsize{2.9} \cellcolor[HTML]{FCE7E4} 
        & 21.8 \textbf{\textcolor[rgb]{0.75,0.0,0.0}{↓}} \scriptsize{2.5} \cellcolor[HTML]{FCE7E4} \\
        \hline
    \end{tabular}
    }

    \label{tab:exps_global_mask}
\end{wraptable}
\noindent \textbf{Global Masking Fine-Tuning.}
Unlike the local MFT above, we further explore MFT in a global setting, where the mask is learned across all eligible trainable layers without manually selecting local layer groups.
Our initial exploration of language-only backbones in Tab.~\ref{tab:exps_global_mask} shows mixed results: global MFT improves LLaMA2-7B in the math domain, especially on GSM8K, but degrades performance on LLaMA3.1-8B.
We note that this weaker performance may be due to the sparsity configuration used in the initial study, where we used around 10\% sparsity to match the local mask-learning setting.

To further demonstrate MFT's generality and transferability, we conduct additional experiments on multimodal LLaVA backbones, as shown in Tab.~\ref{tab:exps_mft_llava}.
This setting evaluates MFT beyond pure language models, incorporating recent LLM backbones into multimodal architectures.
Following the official LLaVA fine-tuning setup, where only the connector and LLM are fine-tuned, we learn a global mask across all corresponding trainable layers, including the connector and all LLM layers.
Therefore, unlike the local MFT experiments, this setting does not rely on manually selected layer groups.
Based on the above observation, we use a lower global sparsity ratio of 1\% for multimodal MFT.
The results show that MFT consistently outperforms Best FFT across all evaluated multimodal benchmarks for both LLaVA variants.
These results suggest that global MFT can be effective when the sparsity level is properly controlled, and that the proposed mask-based post-FFT refinement is transferable from language-only LLMs to multimodal LLMs.

\begin{table}[t]
    \caption{Performance of MFT compared to Best FFT on multi-modal benchmarks across two LLaVA backbones. MFT consistently outperforms Best FFT on all evaluated metrics.}
    \centering
    \renewcommand{\arraystretch}{1.25}%
    \resizebox{1.0\columnwidth}{!}{%
    \begin{tabular}{lcccccc}
        \hline
        \textbf{Method} & \textbf{GQA} & \textbf{MMMU} & \textbf{POPE} & \textbf{MME} & \textbf{SQA} & \textbf{TextVQA} \\
        \hline
        \multicolumn{7}{l}{\textit{LLaVA (CLIP + Qwen2.5-0.5B)}} \\
        \hline
        {Pre-Trained Model} & 5.7 & 22.6 & 49.9 & 93.7 & 3.2 & 9.0 \\
        \hline
        {Best FFT} \cellcolor[HTML]{E6F2FB}
        & 55.7 \cellcolor[HTML]{E6F2FB}
        & 34.2 \cellcolor[HTML]{E6F2FB}
        & 85.3 \cellcolor[HTML]{E6F2FB}
        & 1197.9 \cellcolor[HTML]{E6F2FB}
        & 57.6 \cellcolor[HTML]{E6F2FB}
        & 44.2 \cellcolor[HTML]{E6F2FB} \\
        {MFT} \textbf{(Ours)} \cellcolor[HTML]{FCE7E4}
        & 57.5 \textbf{\textcolor[rgb]{0.0, 0.75, 0.0}{$\uparrow$}} \scriptsize{1.8} \cellcolor[HTML]{FCE7E4}
        & 35.0 \textbf{\textcolor[rgb]{0.0, 0.75, 0.0}{$\uparrow$}} \scriptsize{0.8} \cellcolor[HTML]{FCE7E4}
        & 86.3 \textbf{\textcolor[rgb]{0.0, 0.75, 0.0}{$\uparrow$}} \scriptsize{1.0} \cellcolor[HTML]{FCE7E4}
        & 1250.2 \textbf{\textcolor[rgb]{0.0, 0.75, 0.0}{$\uparrow$}} \scriptsize{52.3} \cellcolor[HTML]{FCE7E4}
        & 58.8 \textbf{\textcolor[rgb]{0.0, 0.75, 0.0}{$\uparrow$}} \scriptsize{1.2} \cellcolor[HTML]{FCE7E4}
        & 46.1 \textbf{\textcolor[rgb]{0.0, 0.75, 0.0}{$\uparrow$}} \scriptsize{1.9} \cellcolor[HTML]{FCE7E4} \\
        \hline
        \multicolumn{7}{l}{\textit{LLaVA (SigLIP + Gemma-2B)}} \\
        \hline
        {Pre-Trained Model} & 2.6 & 26.4 & 56.6 & 586.2 & 2.8 & 5.6 \\
        \hline
        {Best FFT} \cellcolor[HTML]{E6F2FB}
        & 60.9 \cellcolor[HTML]{E6F2FB}
        & 31.3 \cellcolor[HTML]{E6F2FB}
        & 87.4 \cellcolor[HTML]{E6F2FB}
        & 1358.5 \cellcolor[HTML]{E6F2FB}
        & 60.5 \cellcolor[HTML]{E6F2FB}
        & 50.7 \cellcolor[HTML]{E6F2FB} \\
        {MFT} \textbf{(Ours)} \cellcolor[HTML]{FCE7E4}
        & 61.8 \textbf{\textcolor[rgb]{0.0, 0.75, 0.0}{$\uparrow$}} \scriptsize{0.9} \cellcolor[HTML]{FCE7E4}
        & 32.8 \textbf{\textcolor[rgb]{0.0, 0.75, 0.0}{$\uparrow$}} \scriptsize{1.5} \cellcolor[HTML]{FCE7E4}
        & 88.7 \textbf{\textcolor[rgb]{0.0, 0.75, 0.0}{$\uparrow$}} \scriptsize{1.3} \cellcolor[HTML]{FCE7E4}
        & 1379.4 \textbf{\textcolor[rgb]{0.0, 0.75, 0.0}{$\uparrow$}} \scriptsize{20.9} \cellcolor[HTML]{FCE7E4}
        & 61.9 \textbf{\textcolor[rgb]{0.0, 0.75, 0.0}{$\uparrow$}} \scriptsize{1.4} \cellcolor[HTML]{FCE7E4}
        & 51.5 \textbf{\textcolor[rgb]{0.0, 0.75, 0.0}{$\uparrow$}} \scriptsize{0.8} \cellcolor[HTML]{FCE7E4} \\
        \hline
    \end{tabular}
    }
    \label{tab:exps_mft_llava}
\end{table}

\begin{wraptable}{r}{0.42\columnwidth}
    \centering
    \vspace{-20pt}
    \footnotesize
    \caption{
    Comparison of MFT with the best FFT and common regularization techniques.
    }
    \vspace{2pt}
    \renewcommand{\arraystretch}{1.15}

    \resizebox{\linewidth}{!}{%
    \begin{tabular}{lc}
        \hline
        \textbf{Method} & \textbf{MBPP} \\
        \hline
        Zero-Shot & 46.8 \\
        Best FFT & 50.5 \\
        \hspace{1em}+ Dropout & 50.7 \\
        \hspace{1em}+ Stronger Weight Decay & 50.5 \\
        \hspace{1em}+ MFT \textbf{(Ours)} & \textbf{51.2} \\
        \hline
    \end{tabular}
    }

    \label{tab:mbpp_regularization}
    \vspace{-10pt}
\end{wraptable}
\noindent \textbf{Comparison with Regularization Techniques.}
Tab.~\ref{tab:mbpp_regularization} compares MFT with common regularization techniques on MBPP using DeepSeek-Coder-1.3B~\citep{guo2024deepseekcoderlargelanguagemodel}.
Compared with Best FFT, adding dropout yields a slight improvement from 50.5 to 50.7, whereas stronger weight decay does not improve final performance.
In our experiments, dropout provides a short-term gain during continued FFT, but performance soon declines as training proceeds. Stronger weight decay mainly delays the onset of degradation rather than improving the best achievable result.
In contrast, MFT further improves the Best FFT checkpoint to 51.2, outperforming both regularization baselines.
These results suggest that the gain of MFT cannot be simply explained by conventional regularization effects.

Conceptually, MFT is different from classical regularization methods.
Regularization techniques such as dropout, L1/L2 penalties, and weight decay still optimize the model within the full hypothesis space, while constraining the optimization trajectory or parameter magnitude.
MFT instead freezes the well-trained weights and directly optimizes a binary structural mask, thereby selecting a task-adaptive subnetwork from the fine-tuned model.
Therefore, MFT is a structure-level post-FFT refinement method rather than a standard regularization strategy.

\begin{figure*}[t]
    \centering
    \includegraphics[width=1\linewidth]{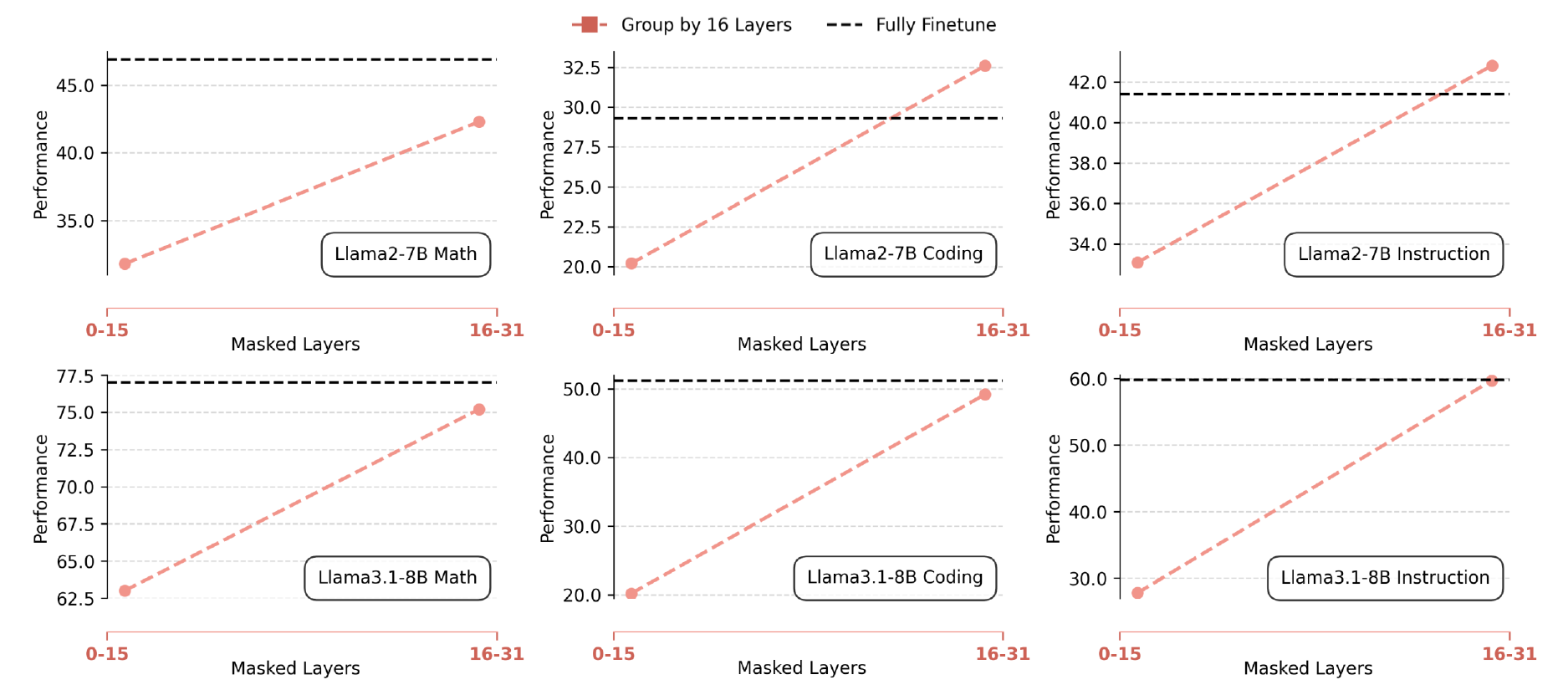}
    \caption{The local MFT ablation results of the larger layer-wise group (16-layer).
    }
    \label{fig:supp_local_ablation_16-layers}
\end{figure*}

\noindent \textbf{Larger Layer-wise Group Ablation.}
The results of larger layer-wise group (16 layers) ablation for local MFT are shown in Fig.~\ref{fig:supp_local_ablation_16-layers}.

\noindent \textbf{Fine-grained Evaluation Details.}
We provide fine-grained evaluation details for FFT and MFT on three domains using LLaMA2-7B in Tab.~\ref{tab:llama2_detailed_eval_domain_specific} (domain-specific FFT) and Tab.~\ref{tab:llama2_detailed_eval_domain_mixed} (domain mixed-up FFT).
They show the epoch number for Best FFT, Continued FFT, Best MFT, and Complete MFT, along with the corresponding token usage.
The number of Continued FFTs is consistently 4, whereas the number for Best FFT is less than 4.
Based on the Best FFT, we then perform 2 epochs of MFT.
Therefore, the number of Complete MFT is always equal to the number of Best FFT plus 2, but Best MFT may be less than it.
For example, in the math domain in Tab.~\ref{tab:llama2_detailed_eval_domain_specific}, Continual FFT runs for 4 epochs, while FFT achieves the best results at 2.4 epochs as Best FFT.
Starting from 2.4 epochs, we perform MFT for an additional 2 epochs. Thus, the Complete MFT is 2.4 + 2 = 4.4 epochs, while the Best MFT is 4 epochs, indicating that MFT uses 4 - 2.4 = 1.6 epochs to achieve the best result based on Best FFT.
For each setting, the number of tokens used is provided in brackets.

\begin{wraptable}{r}{0.46\columnwidth}
    \centering
    \vspace{-16pt}
    \footnotesize
    \caption{
    Exemplar training loss statistics of LLaMA2-7B models after training are stable.
    }
    \vspace{2pt}

    \renewcommand{\arraystretch}{1.05}
    \resizebox{\linewidth}{!}{%
    \begin{tabular}{lccc}
        \toprule
        LLaMA2-7B & Math & Coding & Instruction \\
        \midrule
        FFT loss & 0.101 & 0.098 & 0.125 \\
        MFT loss & 0.085 & 0.054 & 0.060 \\
        \bottomrule
    \end{tabular}
    }

    \label{tab:train-loss}
    \vspace{-16pt}
\end{wraptable}
\noindent \textbf{More Details of Theoretical Analysis.}
For our case, to theoretically support that our MFT has better optimization potential than FFT, we need to verify that $\Delta_{\mathrm{train}} + \Delta_{\mathrm{complexity}} < 0$ in Eq.~\ref{eq:revised}.

For the first term, we first supplement exemplar training-loss statistics in Tab.~\ref{tab:train-loss}. It shows the mean training loss after training stabilizes, indicating that MFT can further reduce training loss compared with the base model (the Best FFT).

Then, we can ensure the first term $\Delta_{\mathrm{train}} < 0$.

For the second term, we need to compare the complexity of the encoded model. Given $d$ as the total number of weights, $z$ as the weights being masked out (as zero), 
and $b$ as the bit width. Since we are using weight-level masking granularity, we have
\begin{equation}\label{eq:cfft}
C(h_{\mathrm{FFT}}) \approx bd, 
\quad 
C(h_{\mathrm{MFT}}) \approx b(1-p)d + \log_2 \binom{d}{z},
\end{equation}
where $p=z/d$ represents the sparse ratio in our case.  
The first term of $C_{\mathrm{MFT}}$ is the encoding cost of non-zero weights, and the second is for the index cost of zero weights.  
Further, we have $\log_2 \binom{d}{z} \approx dH(p)$, where 
\begin{equation}\label{eq:entropy}
H(p) = -p \log_2 p - (1-p)\log_2(1-p).
\end{equation}
Then, the second term can be described as
\begin{equation}\label{eq:dcomplexity}
\Delta_{\mathrm{complexity}} = C(h_{\mathrm{MFT}}) - C(h_{\mathrm{FFT}}) 
\approx d[H(p) - bp].
\end{equation}

For practice, we always use $b=8$ or $16$.
We use a $10\%$ sparsity ratio for our MFT, applied to a 4-layer group, yielding a global sparsity ratio of $2.7\% \sim 3\%$ across two backbones.  
To consider the strict case, we use $b=8$ and a $2.7\%$ ratio in the calculation shown below.  
\begin{equation}\label{eq:hpcalc}
H(p) - bp = H(0.027) - 8 \cdot 0.027 
\approx -0.037 < 0.
\end{equation}
Then, we have the second term $\Delta_{\mathrm{complexity}} < 0$.

Combining both the first and second terms, we have
\begin{equation}\label{eq:final}
U(h_{\mathrm{MFT}}) - U(h_{\mathrm{FFT}})
= \Delta_{\mathrm{train}} + \Delta_{\mathrm{complexity}} < 0.
\end{equation}

Finally, we conclude that $U(h_{\mathrm{MFT}}) < U(h_{\mathrm{FFT}})$, meaning MFT can further reduce the test-loss upper bound compared with FFT and theoretically supports our proposed method.

\begin{figure*}[t]
    \centering
    \includegraphics[width=0.9\linewidth]{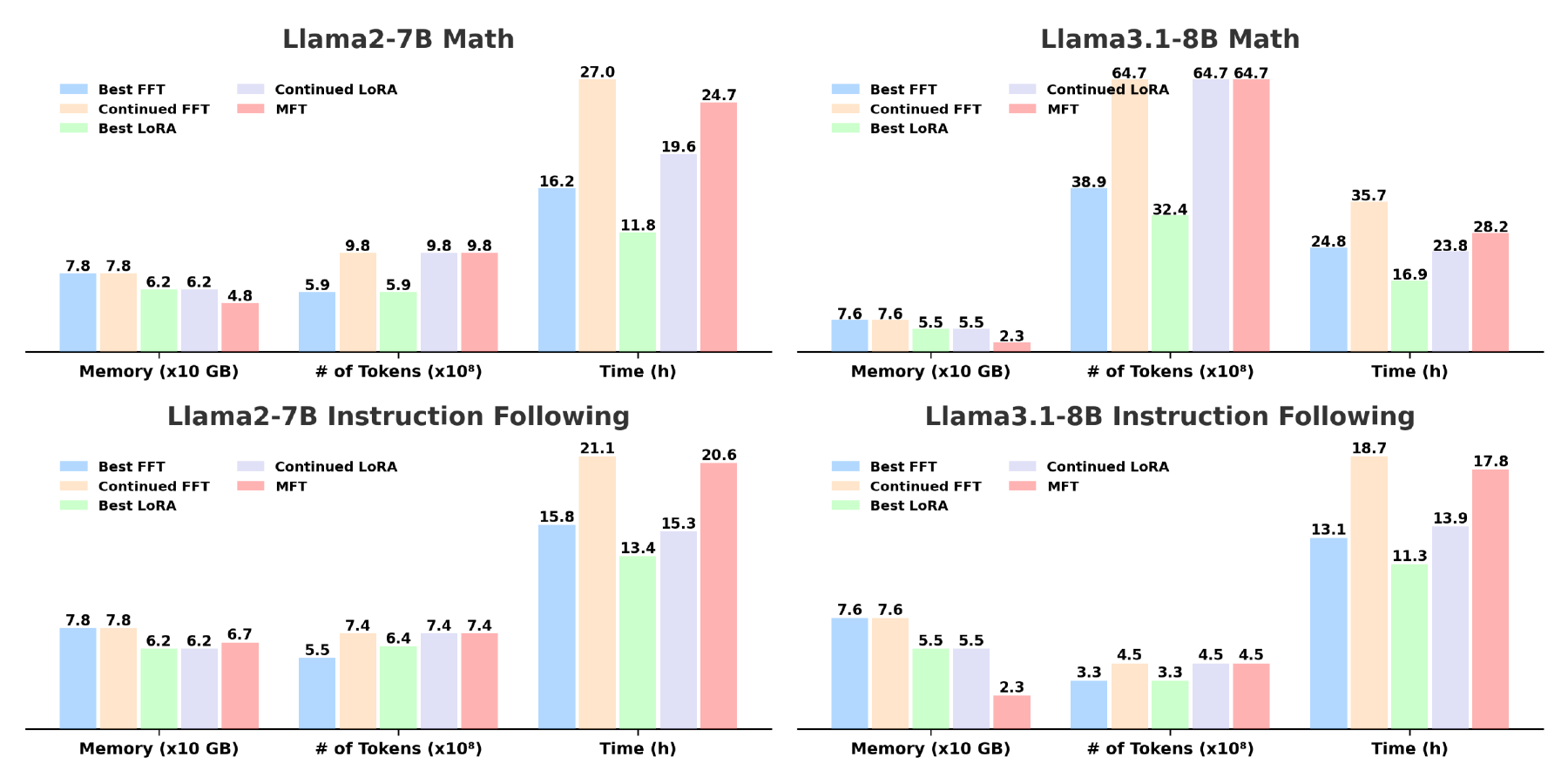}
    \caption{Training cost comparisons of GPU memory, used token number, and training time on math and instruction-following domains using LLaMA2-7B and LLaMA3.1-8B.}
    \label{fig:supp_cost_comparison}
\end{figure*}

\noindent \textbf{More Training Cost Comparisons.}
We supplement the training cost comparisons with GPU memory usage, token count, and training time for the math and instruction-following domains in Fig.~\ref{fig:supp_cost_comparison}.

\begin{table*}[ht]
  \caption{Training epochs and corresponding number of used tokens (in brackets) on LLaMA2-7B under domain‐specific FFT setting.}
  \centering
  \resizebox{\textwidth}{!}{
  \begin{tabular}{lcccc}
    \toprule
    \textbf{Domain} & \textbf{Best FFT} & \textbf{Continued FFT} & \textbf{Best MFT} & \textbf{Complete MFT} \\
    \midrule
    Math          & 2.4 epochs (5.90e8)  & 4.0 epochs (9.83e8)  & 4.0 epochs (9.83e8)  & 4.4 epochs (1.08e9) \\
    Coding    & 2.7 epochs (3.32e8)       & 4.0 epochs (4.92e8)  & 3.6 epochs (4.43e8)  & 4.7 epochs (5.78e8) \\
    Instruction Following & 2.4 epochs (5.51e8)       & 4.0 epochs (9.18e8)  & 3.2 epochs (7.35e8)  & 4.4 epochs (1.01e9) \\
    \bottomrule
  \end{tabular}
  }
  \label{tab:llama2_detailed_eval_domain_specific}
\end{table*}

\begin{table*}[ht]
  \caption{Training epochs and corresponding number of used tokens (in brackets) on LLaMA2-7B under mixed-up FFT setting.}
  \centering
  \resizebox{\textwidth}{!}{
  \begin{tabular}{lcccc}
    \toprule
    \textbf{Domain} & \textbf{Best FFT} & \textbf{Continued FFT} & \textbf{Best MFT} & \textbf{Complete MFT} \\
    \midrule
    Math          & 2.0 epochs (2.67e9)  & 4.0 epochs (5.34e9)  & 3.7 epochs (3.09e9)  & 4.0 epochs (3.16e9) \\
    Coding    & 2.0 epochs (2.67e9)  & 4.0 epochs (5.34e9)  & 3.3 epochs (2.83e9)  & 4.0 epochs (2.92e9) \\
    Instruction Following & 2.0 epochs (2.67e9)  & 4.0 epochs (5.34e9)  & 4.0 epochs (3.13e9)  & 4.0 epochs (3.13e9) \\
    \bottomrule
  \end{tabular}
  }
  \label{tab:llama2_detailed_eval_domain_mixed}
\end{table*}

\begin{figure*}[t]
    \centering
    \includegraphics[width=0.9\linewidth]{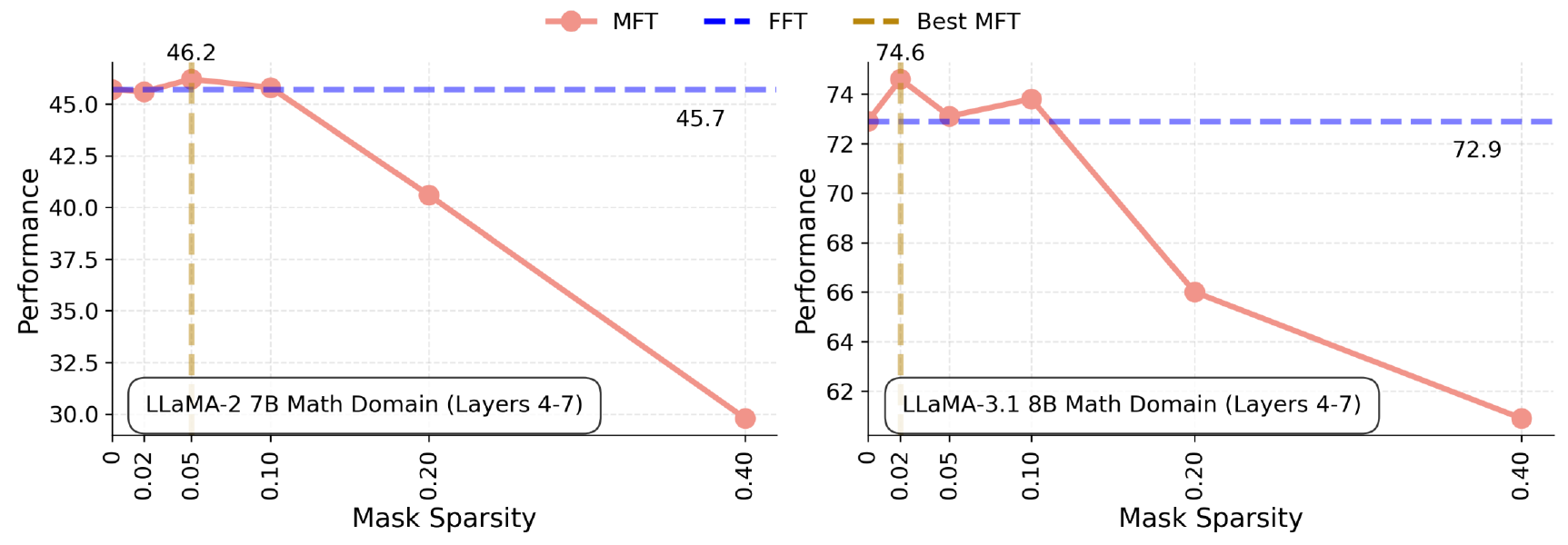}
    \caption{Masking ratio ablation study on the math domain using LLaMA2-7B and LLaMA3.1-8B.
    }
    \label{fig:supp_mask_ratio_ablation_math}
\end{figure*}

\noindent \textbf{More Mask Ratio Ablation.}
The mask ratio ablation on the math domain using LLaMA2-7B and LLaMA3.1-8B is shown in Fig.\ref{fig:supp_mask_ratio_ablation_math}.

\begin{figure*}[t]
    \centering
    \includegraphics[width=0.9\linewidth,height=4.5cm]{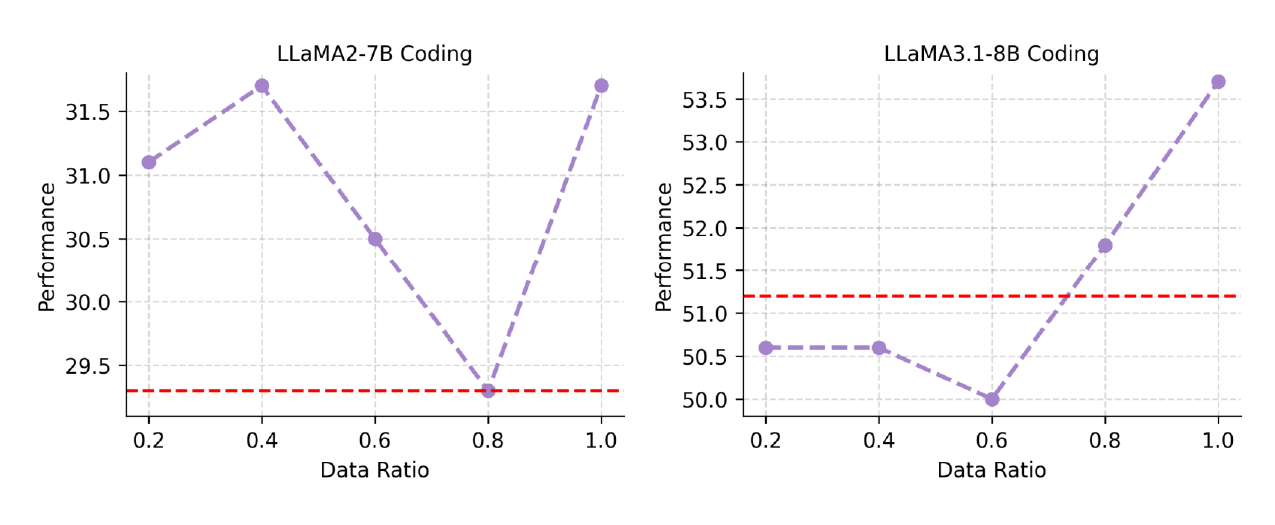}
    \caption{Data ratio ablation study on coding domain using LLaMA2-7B and LLaMA3.1-8B.}
    \label{fig:supp_data_ratio_ablation_coding}
\end{figure*}

\noindent \textbf{More Data Ratio Ablation.}
The MFT data ratio ablation on the coding domain using LLaMA2-7B and LLaMA3.1-8B is shown in Fig.~\ref{fig:supp_data_ratio_ablation_coding}.

\begin{figure*}[t]
    \centering
    \includegraphics[width=0.9\linewidth]{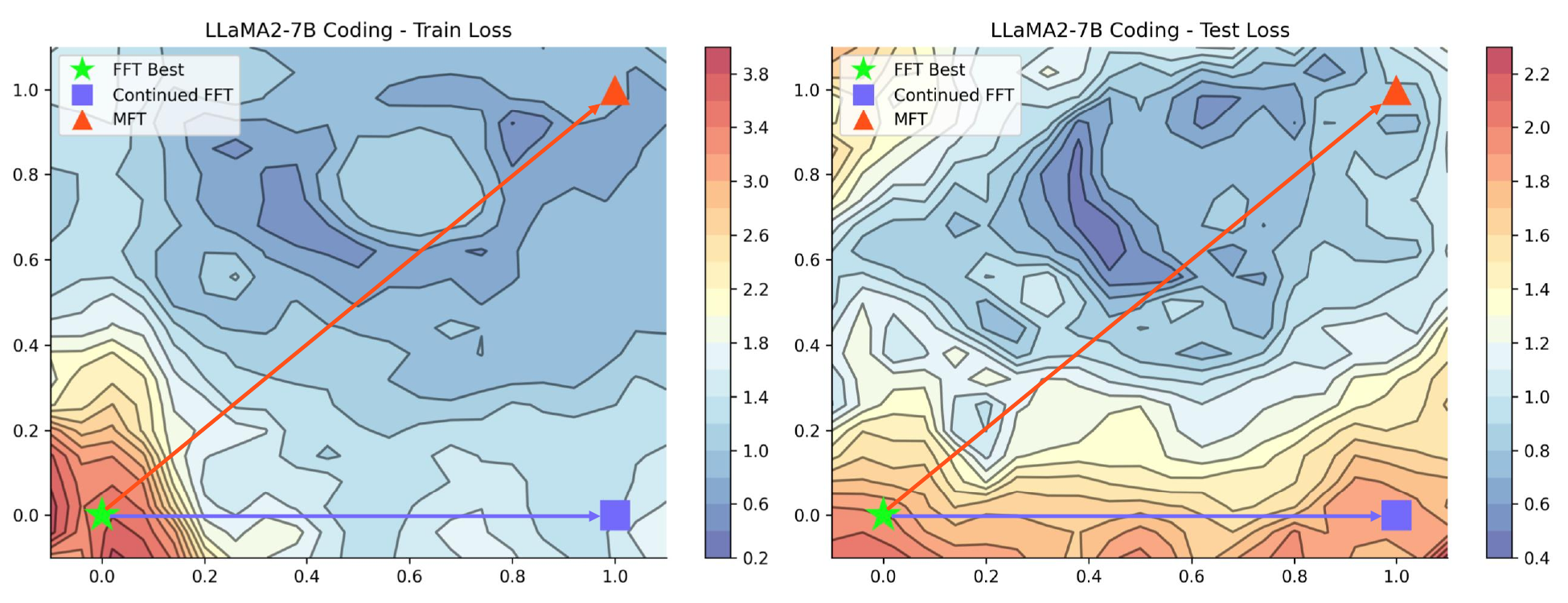}
    \caption{The loss landscape visualization on the coding domain using LLaMA2-7B.
    }
    \label{fig:supp_loss_surface_coding}
\end{figure*}

\noindent \textbf{More Details of Loss Landscape Visualization.}
A loss landscape visualization in the coding domain using LLaMA2-7B is shown in Fig.~\ref{fig:supp_loss_surface_coding}.

For the loss surface analysis, we present empirical results as loss-landscape visualizations in Fig.~\ref{fig:loss_surface} and Fig.~\ref{fig:supp_loss_surface_coding}. Here, we provide further support for the theoretical view.

We assume the model after FFT (Best FFT) has Hessian as 
\begin{equation}\label{eq:hfft}
H_{\mathrm{FFT}} = \nabla^2 L(\theta_{\mathrm{FFT}}) \succeq 0.
\end{equation}

Since MFT learns a mask and applies it to the model after FFT, which means the MFT model is a projection $P$ of the FFT model, where the corresponding Hessian can be given by
\begin{equation}\label{eq:hmft}
H_{\mathrm{MFT}} = P H_{\mathrm{FFT}} P.
\end{equation}

Since for $H_{\mathrm{FFT}} \succeq 0$ and its any projection ($H_{\mathrm{MFT}}$ for here), we have
\begin{equation}\label{eq:eigs}
\begin{aligned}
&\lambda_{\max}(H_{\mathrm{MFT}}) \leq \lambda_{\max}(H_{\mathrm{FFT}}), \\
&\mathrm{Tr}(H_{\mathrm{MFT}}) \leq \mathrm{Tr}(H_{\mathrm{FFT}}).
\end{aligned}
\end{equation}

As $\lambda_{\max}$ and $\mathrm{Tr}(\cdot)$ represent model sharpness and mean curvature, lower values indicate a flatter model, meaning the MFT model generalizes better than the FFT model. 
Such analysis provides theoretical support and aligns with the empirical results shown in the landscape visualizations (the MFT model is visually flatter than the FFT model).

\section{More Discussion}\label{sec:conclusion-supp}
\noindent \textbf{Intuition.}
MFT is highly relevant to a range of well-established research topics, such as sparse network training and network pruning.
However, inspired by their techniques, we approach sparsity from a novel perspective, shifting from typical model efficiency to a general capability perspective by removing negative components to improve performance.
In other words, sparse models typically aim to compress the model while maintaining performance (often referred to as \textit{subtraction}). However, we use sparsity as a tool for model enhancement (seen as \textit{addition}).
We expect that such counterthinking about sparse networks could inspire broader explorations in the future, enriching their research potential.


\end{document}